%% file: main-7125-Jiang.tex
\documentclass[11pt,a4paper]{article}
\usepackage{times,latexsym}
\usepackage{url}
\usepackage[T1]{fontenc}

\input{def}
\input{std-macros}

\usepackage[acceptedWithA]{tacl2021v1}

\usepackage{xspace,mfirstuc,tabulary}

\newif\iftaclinstructions
\taclinstructionsfalse 
\iftaclinstructions
\renewcommand{\confidential}{}
\renewcommand{\anonsubtext}{(No author info supplied here, for consistency with
TACL-submission anonymization requirements)}
\newcommand{\instr}
\fi

\iftaclpubformat 

\else

\fi

\title{Few-Shot Multilingual Open-Domain QA from 5 Examples}

\author{Fan Jiang \and Tom Drummond \and Trevor Cohn\Thanks{Also at Google.} \\
  School of Computing and Information Systems \\
  The University of Melbourne, Victoria, Australia \\
  \texttt{fan.jiang1@student.unimelb.edu.au}\\
  \texttt{\{tom.drummond, trevor.cohn\}@unimelb.edu.au}}

\date{}

\begin{document}
\maketitle
\begin{abstract} 
Recent approaches to multilingual open-domain question answering (MLODQA) have achieved promising results given abundant language-specific training data.
However, the considerable annotation cost limits the application of these methods for underrepresented languages.
We introduce a \emph{few-shot learning} approach to synthesise large-scale multilingual data from large language models (LLMs).
Our method begins with large-scale self-supervised pre-training using WikiData, followed by training on high-quality synthetic multilingual data generated by prompting LLMs with few-shot supervision.
The final model, \ours, significantly outperforms existing few-shot and supervised baselines in MLODQA and cross-lingual and monolingual retrieval.
We further show our method can be extended for effective zero-shot adaptation to new languages through a \emph{cross-lingual prompting} strategy with only English-supervised data, making it a general and applicable solution for MLODQA tasks without costly large-scale annotation.
\end{abstract}

\input{sections/1_introduction}
\input{sections/2_method}
\input{sections/3_experiments}
\input{sections/4_relatedwork}

\section{Conclusion and Limitation}
In this work, we propose \ours, a \emph{few-shot learning} approach for multilingual open-domain retrieval tasks. We present a novel self-supervised pre-training framework that exploits WikiData to effectively initialise both multilingual retrieval and QA capabilities. This process is followed by few-shot synthetic multilingual QA generation from LLMs using only five human-annotated examples. We demonstrate that the resulting model achieves competitive multilingual retrieval and QA performance through fine-tuning on the high-quality synthetic data. We further show that this few-shot approach generalises to zero-shot settings that only require English-supervised data. This mechanism serves as an effective approach for language adaptation, enabling the adapted model to achieve both boosted retrieval and end-to-end QA performance across fifteen previously unseen languages.

This work uses LLMs for synthetic data generation, which may propagate undesirable biases to generated data. We believe such biases will not be amplified as we sample prompts from \textsc{Xor-TyDi QA}, a dataset annotated with strict guidelines. Our preliminary safety analysis also reveals that only less than 1\% data contains potentially harmful queries, as identified by \texttt{Llama-Guard-2}.

\section*{Acknowledgement}
We thank the action editor Shay Cohen and anonymous reviewers for their helpful feedback and suggestions. The first author is supported by the Graduate Research Scholarships funded by the University of Melbourne. This work was funded by the Australian Research Council, Discovery grant DP230102775.

\bibliography{tacl2021}
\bibliographystyle{acl_natbib}

\clearpage

\appendix
\input{sections/appendix}

\end{document}

%% file: def.tex
\usepackage{multirow}
\usepackage{makecell}
\usepackage{hhline}
\usepackage{booktabs}
\usepackage{bm}
\usepackage{amssymb}
\usepackage{amsmath}
\usepackage{url}
\usepackage{graphicx}
\usepackage{float}
\usepackage{subfigure}
\usepackage{subcaption}
\usepackage{paralist}
\usepackage[ruled, lined, linesnumbered, commentsnumbered, longend]{algorithm2e}
\usepackage{xcolor}
\usepackage{colortbl}
\usepackage{pifont}
\usepackage{tabularx}
\usepackage[T2A,LAE,T1]{fontenc}
\usepackage{CJKutf8}
\usepackage{enumitem}
\setlist[itemize]{noitemsep, topsep=0pt}
\usepackage[russian,english]{babel}
\usepackage{relsize}

\DeclareRobustCommand{\cyrins}[1]{
  \begingroup\fontfamily{cmr}
  \foreignlanguage{russian}{#1}
  \endgroup
}
\DeclareRobustCommand{\aratext}[1]{
  \begingroup\fontfamily{Arab}
  \foreignlanguage{arabic}{#1}
  \endgroup
}

\definecolor{LightCyan}{rgb}{0.88,1,1}

\newcommand{\promptsize}{\fontsize{7pt}{8pt}\selectfont}
\newcommand{\propertytsize}{\fontsize{7pt}{8pt}\selectfont}
 
\newcommand{\class}{CLASS\xspace}

\newcommand{\ours}{\textsc{FsModQA}\xspace}
\newcommand{\oursen}{\textsc{FsModQA-En}\xspace}
\newcommand{\ourdata}{\textsc{FsMlQA}\xspace}
\newcommand{\ourptdata}{\textsc{MlWikiQA}\xspace}

\usepackage{ntheorem}

\theoremstyle{nonumberplain}

\def\eg{{e.g.,}\xspace}
\def\ie{{i.e.,}\xspace}
\def\versus{{\em v.s.}\xspace}

\definecolor{lightblue}{HTML}{bdd6fb}
\definecolor{boxgray}{gray}{0.9}
\definecolor{bgyellow}{HTML}{fcebde}
\definecolor{bgred}{HTML}{d77470}
\definecolor{bggrey}{HTML}{dcc0e5}

\usepackage{inconsolata}
\usepackage{pifont}
\usepackage[most]{tcolorbox}
\usepackage{csquotes}
\usepackage{anyfontsize}
\usepackage{comment}
\usepackage{float}
\usepackage{cuted}
\usepackage{tikz}
\usepackage{listings,multicol}
\usepackage{etoolbox}
\BeforeBeginEnvironment{lstlisting}{\begin{mdframed}\vspace{-0.7em}}
\AfterEndEnvironment{lstlisting}{\vspace{-0.5em}\end{mdframed}}

\def\ifempty#1{\def\temparg{#1}\ifx\temparg\empty}

\usepackage{caption}

\newtcolorbox[list inside=prompt,auto counter,number within=section]{prompt}[1][]{
    colbacktitle=black!60,
    coltitle=white,
    colback=bgyellow,
    fontupper=\footnotesize,
    boxsep=5pt,
    left=0pt,
    right=0pt,
    top=0pt,
    bottom=0pt,
    boxrule=1pt,
    #1,
}

\def\bng{\bngx}

\font\bngx=bang10

\def\*#1*#2{o\null{#2}{#1}}


%% file: std-macros.tex
\newcommand\ba{\ensuremath{\bm{a}}}

\newcommand\bd{\ensuremath{\bm{d}}}

\newcommand\bq{\ensuremath{\bm{q}}}

\newcommand\bx{\ensuremath{\bm{x}}}

\usepackage{color}


%% file: sections/1_introduction.tex
\section{Introduction}

\begin{figure*}
    \setlength{\belowcaptionskip}{-0.35cm}
    \centering
    \includegraphics[width=0.8\linewidth]{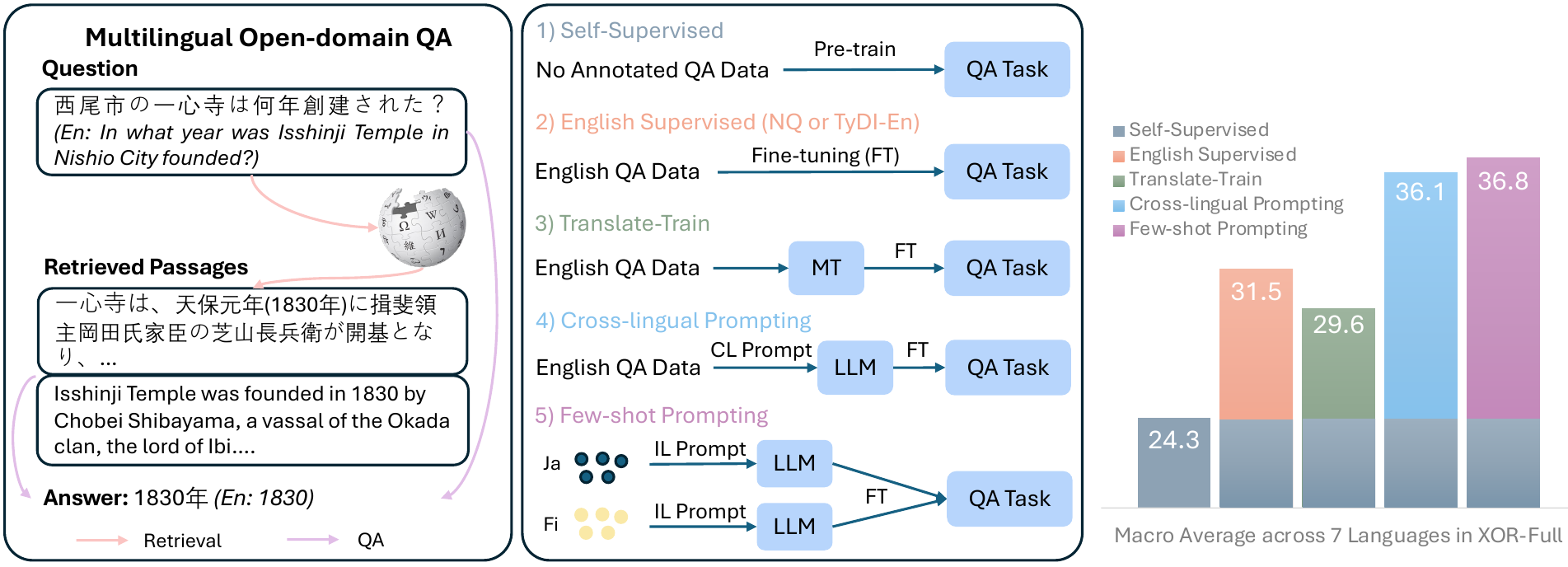}
    \caption{\textbf{Left (a)}: The process of multilingual open-domain QA. \textbf{Middle (b)}: training strategies: 1) self-supervised pre-training; 2) fine-tuning on English QA; 3) translate English QA to target languages; 4) use English data to prompt LLMs to generate target language data; 5) use few-shot in-language data for LLM prompting. \textbf{Right (c)}: Performance comparison (Avg. F1) on the XOR-Full dataset.}
    \label{fig:overview}
\end{figure*}

Open-domain QA has demonstrated impressive performance by employing the \emph{retrieve-then-read} (Figure~\ref{fig:overview}(a)) pipeline~\citep{chen-etal-2017-reading}, which is built upon dense retrievers~\citep{karpukhin-etal-2020-dense} and efficient generative readers~\citep{izacard-grave-2021-leveraging}. However, this success has been primarily limited to English, leaving the multilingual setting under-explored. This limitation is mainly due to the difficulty and costs of creating high-quality and balanced human-supervised training data for languages other than English. Moreover, multilingual open-domain QA introduces additional challenges with retrieving evidence from multilingual corpora, requiring the underlying retrieval system to be capable of both cross-lingual and monolingual retrieval~\citep{asai2021one}.

More recently, efforts have been made to create multilingual open-domain QA benchmarks from existing multilingual machine reading comprehension tasks (\eg \textsc{Xor-TyDi QA}~\citep{asai-etal-2021-xor}) and by translating English datasets (\eg MKQA~\citep{longpre-etal-2021-mkqa}). These datasets have enabled various approaches to address multilingual open-domain QA problems, including iterative data augmentation~\citep{asai2021one} and extensive additional pre-training on Wikipedia texts~\citep{lapca, jiang-etal-2024-pre}. However, these methods still heavily depend on abundant high-quality language-specific data for fine-tuning, making them less effective solutions when language resources are limited. Therefore, a more generalisable approach to multilingual open-domain QA should aim to mitigate this reliance and be capable of facilitating language adaptation with minimally supervised samples.

In this paper, we present {\bf\ours}, a method for \emph{{\bf\emph{F}}ew-{\bf\emph{S}}hot} {\bf M}ultilingual {\bf O}pen-{\bf D}omain {\bf QA} using minimally-sized supervised data (\ie up to 5 per language).\footnote{We use the term \emph{few-shot} throughout this paper to denote that our method relies on only a small number of human-annotated examples. Thus, we classify our method as a \emph{few-shot learning} approach, consistent with~\citet{dai2023promptagator}.} Our approach consists of two core components: a self-supervised pre-training objective on multilingual corpora; and a synthetic data generation pipeline that prompts a large language model (LLM) using few-shot supervised examples. Concretely, we generate question-answer pairs from WikiData triples by leveraging LLMs' In-Context Learning (ICL) ability. To facilitate ICL prompts, we incorporate ChatBots to generate curated input-output pairs, which serve as examples for prompting LLMs to generate millions of questions from WikiData triples across various languages. After generating these question-answer pairs, we identify the supported Wikipedia passages through answer string matching. We further gather cross-lingual answers and evidence passages through Wikipedia language links to facilitate cross-lingual retrieval. Employing this generated data, we train a multilingual model with a joint objective for retrieval and QA, producing a promising pre-trained model (Figure~\ref{fig:overview}(c)) for subsequent \emph{few-shot learning}.

In \emph{few-shot learning}, we employ LLMs for data generation from few-shot examples. For each target language, we feed the few-shot examples to an LLM and prompt it to generate question-answer pairs from a given document. The few-shot examples are assumed to encapsulate the QA style and distribution of the target dataset, enforcing the LLM to generate synthetic data with similar characteristics. With abundant data, the pre-trained model can be further fine-tuned to achieve superior results (Figure~\ref{fig:overview}(c)). As an unsupervised alternative, we explore a \emph{zero-shot cross-lingual prompting} strategy that uses data from other languages as prompts for data generation, and we show this is almost on par with \emph{few-shot prompting} (Figure~\ref{fig:overview}(c)). 

We evaluate \ours on various datasets, including cross-lingual and monolingual retrieval, and multilingual open-domain QA. We observe notable improvements over competitive few-shot baselines, with +5.1\% gain on retrieval and +8.4\% gain on multilingual open-domain QA. To further test \ours's language adaptation ability, we conduct zero-shot adaptation experiments using our \emph{cross-lingual prompting} strategy on fifteen languages. This adaptation improves performance in both monolingual retrieval and multilingual QA significantly, achieving results that are superior or comparable to strong translation-based methods.\footnote{Code, data, and checkpoints are available \href{https://github.com/Fantabulous-J/FSMODQA}{here}.}

%% file: sections/2_method.tex
\section{\ours}

\begin{figure*}
    \setlength{\belowcaptionskip}{-0.3cm}
    \centering
    \includegraphics[width=0.9\linewidth]{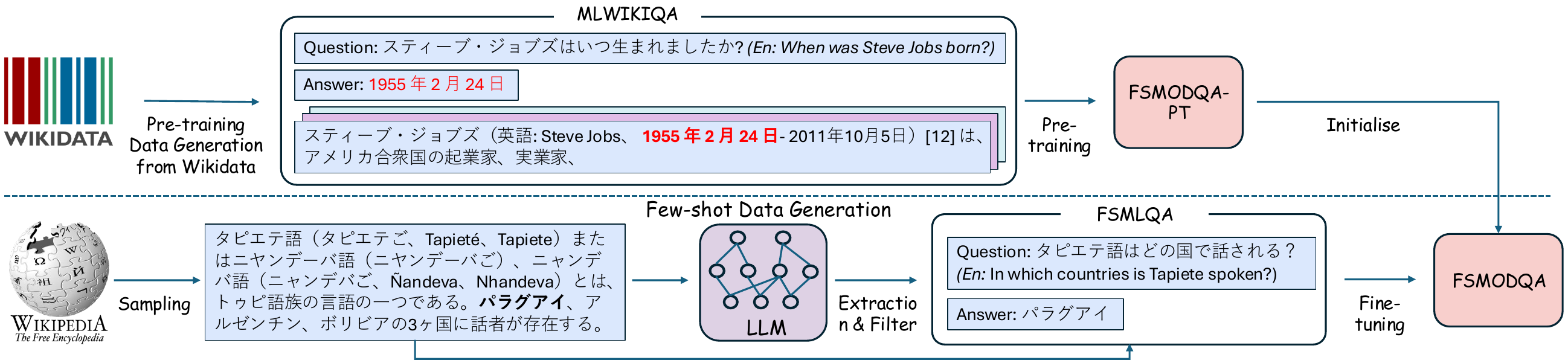}
    \caption{Full pipeline for data construction and model training: (1) generate large-scale data from Wikidata for self-supervised pre-training; (2) use few-shot prompting to generate synthetic Q\&A pairs from Wikipedia passages of target languages, on which the pre-trained model is further fine-tuned.}
    \label{fig:fsmodqa_pipeline}
\end{figure*}

Figure~\ref{fig:fsmodqa_pipeline} presents the full pipeline for generating self-supervised pre-training and fine-tuning data.
\subsection{Self-Supervised Data Construction}\label{sec: mlwikiqa}

\paragraph{Sampling Factual Triplets.}
Our self-supervised training dataset is constructed based on Wikidata~\citep{wikidata}, a multilingual knowledge base consisting of fact triplets linked to millions of entities. We manually select 50 common properties (Appendix Table~\ref{tab:wikidata_property}) based on English and consider all triples associated with these relations. We then gather fact triplets in the desired target languages through language links.

\paragraph{Generating Questions.}
Given a triplet $\mathcal{T}=(s,r,o)$, we aim to write a question $q$ about the head entity $s$'s property $r$ with the gold answer $a$ being the tail entity $o$. One can use relation-specific templates to efficiently transform each triple into natural questions~\citep{sciavolino-etal-2021-simple}. However, this method lacks diversity, making triples with the same properties generate questions with similar surface forms. Instead, we adopt a generative approach by using a LLM to automatically generate questions with more diverse styles.

Specifically, we first sample five triples for each property and prompt ChatGPT (\texttt{gpt-3.5-turbo}) to generate three questions for each triple. This process yields a curated set of high-quality questions: $\mathbb{K}=\{s_i, r_i, o_i, q_i\}_{i=0}^k$.

We additionally generate questions with \texttt{Yes/No} answers from the same set of sampled triples. It is easy to generate \texttt{Yes} questions. For \texttt{No} questions, we need to create false fact triples from existing triples. Specifically, we randomly replace a triple's head or tail entity with the most similar Wikidata entity, and check the perturbed triple is not a valid fact according to Wikidata. We then generate questions using ChatGPT as before. Examples are included in Appendix Table~\ref{tab:chatgpt_prompt}.

Subsequently, these curated questions are used as in-context learning (ICL) examples to prompt a smaller LLM to transform all sampled triples into natural questions. We use \href{https://huggingface.co/google/gemma-7b}{Gemma-7B}~\citep{team2024gemma} as the LLM and include the prompts we used in Appendix Table~\ref{tab:icl_prompt}.

\begin{figure*}
    \setlength{\belowcaptionskip}{-0.2cm}
    \centering
    \includegraphics[width=0.75\linewidth]{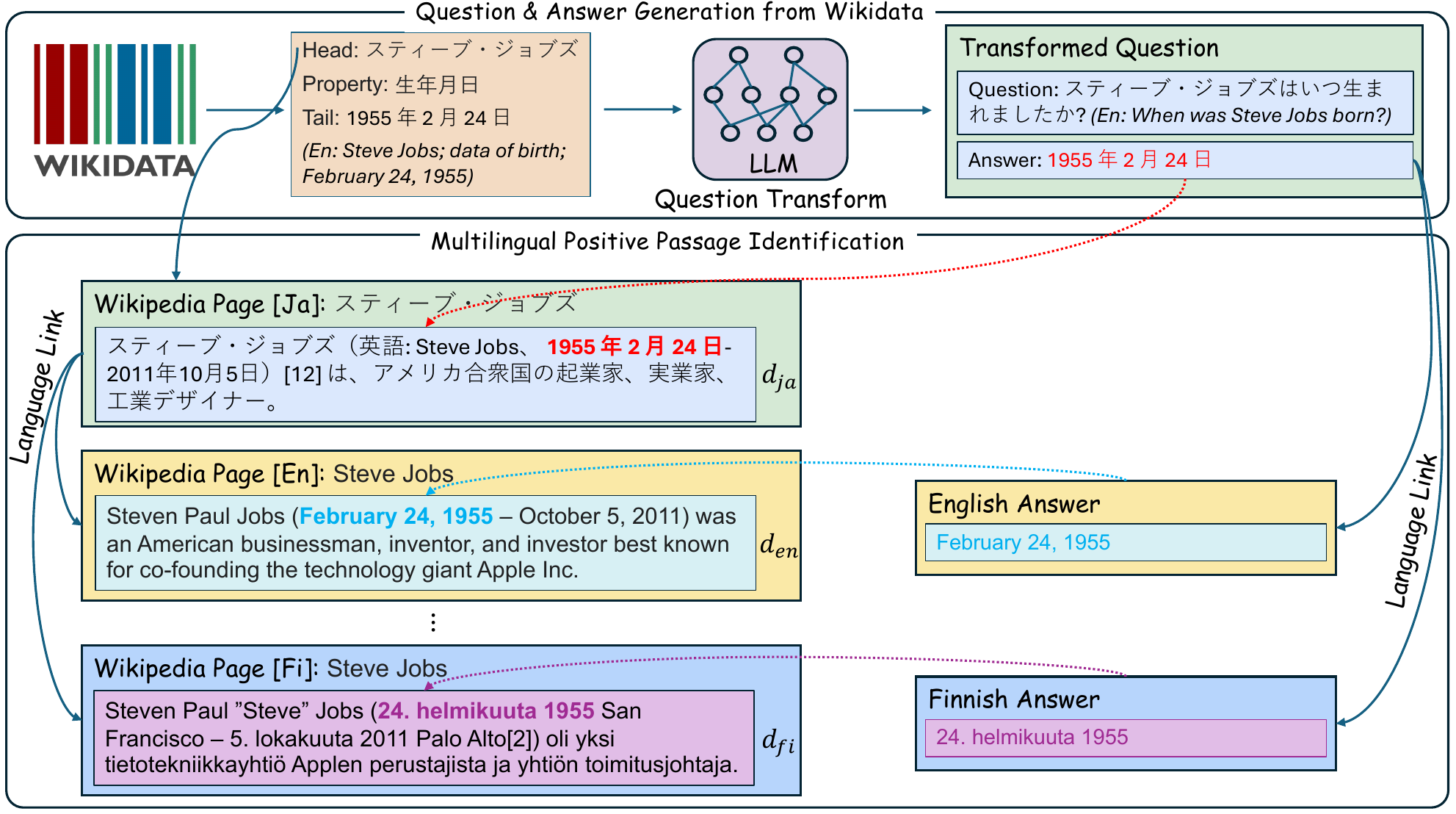}
    \caption{Pre-training data construction pipeline: (1) transform WikiData triples into QAs using LLMs for each target language $L$, and (2) identify in-language and cross-lingual positive passages from the head entity's Wikipedia page and through language links. \emph{English translations} are added for readability.}
    \label{fig:wikidata_pt}
\end{figure*}

\paragraph{Multilingual Positive Passage Identification.}

As shown in Figure~\ref{fig:wikidata_pt}, for a question $q^{ja}$ and answer $a^{ja}$ derived from a triple $(s^{ja},r,o^{ja})$, we gather all passages from the Wikipedia page $\mathcal{W}^{ja}$ linked by $s^{ja}$ and add passages containing $a^{ja}$ as positive $\mathcal{D}_q^{ja}$. If no such passage exists, we use partial match and select the one with the highest lexical overlaps with $a^{ja}$ as positive. We further include positive passages in other languages to facilitate cross-lingual retrieval. We first translate the triple into target languages $(s^L,r,a^L)$ using language links and identify cross-lingual positives by searching $a^L$ in the Wikipedia page $\mathcal{W}^L$ linked by $s^L$ as above. This derives monolingual and cross-lingual positive passages $\mathcal{D}_q=\mathcal{D}_q^{ja}\cup\mathcal{D}_q^{L}$. We generate 18.7M $(q,a,\mathcal{D}_q)$ triples across 8 languages in total, denoted as \ourptdata.\footnote{We classify \ourptdata as a silver-standard dataset rather than a synthetic one, as it is derived from the structured information in WikiData and Wikipedia.}

\subsection{Few-shot Synthetic Data Generation}\label{sec:fs_data_gen}
\paragraph{Few-shot Setting.}
The main idea of \ours is to amplify a limited number of annotated examples into a substantially larger volume of synthetic data by prompting LLMs. In this work, we consider \textsc{Xor-TyDi QA}~\citep{asai-etal-2021-xor} as our target dataset. For each language in \textsc{Xor-TyDi QA}, we randomly sample five triples $\mathcal{K}=\{(q_i^L,a_i^L,d_i^L)\}_{i=1}^5$ from the training set as \emph{few-shot} examples. Each triple contains the question, answer, and the ground truth passage. We ensure that three examples are span answers, while the remaining two are \texttt{Yes} and \texttt{No} answers to align with \textsc{Xor-TyDi QA} distribution.

\paragraph{Prompt-based Question \& Answer Generation.}
We populate a hand-engineered template with our \emph{few-shot} language-specific examples $\mathcal{K}$ and use them as the ICL examples to prompt LLM. Given a randomly sampled passage $d^L$ from language $L$, we append $d^L$ to the template, and the LLM is expected to generate a relevant question $q^L$ and answer $a^L$ in language $L$. We further constrain the answer $a^L$ to be a span within $d^L$, a property of the original \textsc{Xor-TyDi QA} dataset.

Many questions classified as \emph{unanswerable} in~\citet{clark-etal-2020-tydi} can be answered by referring to English Wikipedia~\citep{asai-etal-2021-xor}. These questions are included as \emph{cross-lingual} questions in \textsc{Xor-TyDi QA}. To simulate this scenario, we generate synthetic cross-lingual data from English passages. We first use \href{https://translate.google.com/}{Google Translate} to translate the \emph{few-shot} examples to English: $\mathcal{K}'=\{(q_i^{En}, q_i^{L}, a_i^{En}, a_i^L, d_i^{En})\}_{i=1}^5$. Subsequently, we use these translated \emph{few-shot} examples $\mathcal{K}'$ to fill another template and instruct the LLM to generate QA from a randomly sampled English passage $d^{En}$, first in English $(q^{En}, a^{En})$ and then in target language $(q^{L}, a^{L})$. Similarly, we restrict $a^{En}$ to be a span within $d^{En}$. We include the prompts we used in Tables~\ref{tab:il_prompt} and~\ref{tab:cl_prompt} in Appendix.

\paragraph{Data Filtering.}
We employ a method based on Natural Language Inference (NLI) to enhance the quality of our synthetic data. NLI techniques aim to classify whether a hypothesis text is entailed by, neutral, or contradictory to a given premise text~\citep{bowman-etal-2015-large}. They have been widely used for identifying factual errors in text summarisation~\citep{laban-etal-2022-summac} and hallucinations in machine-generated texts~\citep{honovich-etal-2022-true}. In this study, we employ NLI methods for data filtering~\citep{yoran2024making}. Given a synthetic example $(q,a,d)$, we consider the source passage $d$ as the premise and the concatenation of the generated question $q$ and answer $a$ as the hypothesis. We retain an example only when the premise entails the hypothesis.

In more detail, we apply a novel \emph{local-to-global} filtering mechanism. In \emph{local filtering}, we evaluate whether the originating passage $d$ entails the synthetic QA $(q,a)$ pairs. We take the output probability of the entailment label as the score and keep examples when the entailment score exceeds a threshold $\mathcal{T}_l$. In \emph{global filtering}, we use a pre-trained model (\ie the self-supervised model in Figure~\ref{fig:overview} (b)) to perform retrieval for the question $q$ and obtain a set of passages $\hat{\mathcal{D}}_q$. We compute an entailment score vector $\bx\in\mathcal{R}^{|\hat{\mathcal{D}}_q|}$, with each entry being the entailment score between $(q,a)$ and a retrieved passage $d\in\hat{\mathcal{D}}_q$. We then apply a maximum pooling operation $\operatorname{max}(\bx)$ to derive the final score. The intuition behind this is that a valid $(q,a)$ should be supported by at least one of the retrieved passages, which aligns with open-domain settings. Similarly, we retain only those examples whose scores surpass a predefined threshold $\mathcal{T}_g$. In this way, we end up having 1.7M synthetic data in total across 7 languages, denoted as \ourdata.

\subsubsection{Zero-shot Cross-lingual Prompting}\label{sec:zero_shot_prompting}
Our \emph{few-shot} setting relies on a few annotated examples to generate synthetic QA pairs in target languages. However, this approach encounters significant challenges when the target language is extremely low-resourced, making it nearly impossible to obtain even a few examples. For this setting, we explore \emph{zero-shot} prompting, which uses \emph{cross-lingual} examples to prompt LLMs to generate synthetic QA pairs in target languages.

We consider two \emph{zero-shot} prompting settings. In \emph{English-Prompting} setting, we use English QA data to fill up a template and use it as the prompt to ask LLMs to generate QA pairs from passages randomly sampled from the target language. In \emph{Multilingual-Prompting} setting, we assume access to a handful of examples in a held-out language set. We randomly sample five multilingual examples from this held-out set to populate another template, and prompt LLMs to generate QA pairs in target languages. We include the prompts used in Tables~\ref{tab:en_prompt} and~\ref{tab:multilingual_prompt} in Appendix.

\subsubsection{Data Sampling}\label{sec:geometric_sampling}
Our synthetic dataset, \ourdata, exhibits a strongly skewed distribution towards shorter answer lengths (often single tokens), whereas the human-annotated answers in \textsc{Xor-TyDi QA} tend to be substantially longer. To address this mismatch, we resample the training data from \ourdata according to answer length, using a geometric distribution, $l \sim \operatorname{Geo}(p)$, to achieve a better balance between short and long answers.\footnote{Empirically, we set $p=0.4\ (\mu = 2.5)$ for all languages except for Japanese, where we set $p=0.1\ (\mu = 10)$ to favour longer answers. When computing the distribution, we truncate the answer length to 30.}

\begin{figure}
    \setlength{\belowcaptionskip}{-0.3cm}
    \centering
    \includegraphics[width=\linewidth]{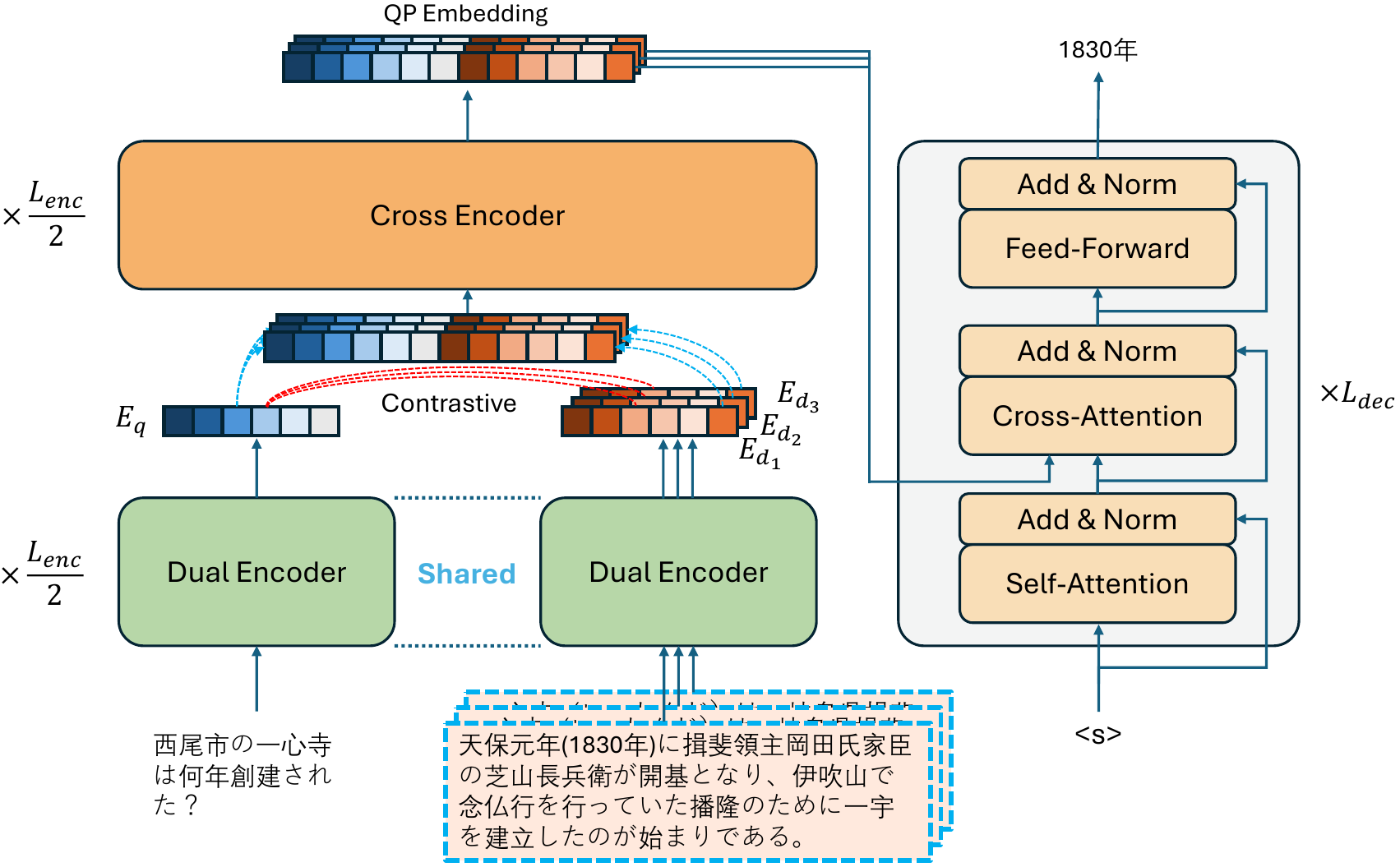}
    \caption{The unified model for passage retrieval and question answering.}
    \label{fig:unified_model}
\end{figure}

\subsection{\ours Model}

\paragraph{Model Structure.}
As shown in Figure~\ref{fig:unified_model}, we employ a single encoder-decoder model to perform both passage retrieval and QA tasks. The first half of the encoder functions as a dual-encoder with shared parameters, which separately encodes the question $\bq$ and the passage corpus $\mathcal{D}$. Additionally, we append an instruction to the question to inform the language of the target answer: "\emph{Answer in $\{\text{lang}\}$}". A LayerNorm operation, followed by average pooling, is applied to compress the inputs into single vectors: $E_{\bq}$ and $\{E_{{\bd}_i}|{\bd}_i\in\mathcal{D}\}$, which are used for matching via dot products. The top-$k$ most relevant passages to the question are selected: $\mathcal{D}_{\bq}=\arg\operatorname{topk}_{{{\bd}_i}\in\mathcal{D}}(E_{\bq}\cdot E_{{\bd}_i})$. The embeddings of the question and each top-$k$ passage in $\mathcal{D}_{\bq}$ are concatenated and fed into the remaining cross-encoder layers. Finally, the cross-encoder embeddings are flattened and incorporated into the decoder through cross-attention to generate the answer ${\ba}$, following the Fusion-in-Decoder approach~\citep{izacard-grave-2021-leveraging}.

\paragraph{Model Training.}
\ours is first pre-trained on \ourptdata and later fine-tuned on \ourdata. In self-supervised pre-training, we use a simple contrastive loss and answer generation loss to train \ours. The dual-encoder is updated by contrasting the paired question passage against the targets of other questions in one training batch (\ie in-batch negative). Formally, for $i$-th training example, the loss function $\mathcal{L}_{\text{ssl}}^i$ is:
\begin{gather}
    \scalebox{1.}{$-\log\frac{e^{(E_{{\bq}_i}\cdot E_{{\bd}_i})}}{\sum_{j=1}^{N}e^{(E_{{\bq}_i}\cdot E_{{\bd}_j})}} - \log \prod_{t=1}^T P({\ba}^i_t|{\ba}^i_{<t},{\bq}_i, {\bd}_i) \nonumber$}
\end{gather}

The pre-trained \ours is subsequently fine-tuned on our synthetic data through an end-to-end training mechanism. The dual-encoder is trained using signals derived from the answer generation task, with the cross-attention score from the decoder serving as the target for assessing question-passage relevance. For $i$-th training example, the loss function is formally defined as:
\begin{gather}
    \scalebox{0.875}{$\mathcal{L}^i_{\text{ret}}=\mathbb{KL}(P_{\text{ret}}(\cdot|{\bq}_i,\mathcal{D}_{{\bq}_i}||P_{\text{ca}}(\cdot|{\bq}_i,\mathcal{D}_{{\bq}_i})),\nonumber$} \\
    \scalebox{0.875}{$P_{\text{ret}}(\cdot|{\bq}_i,\mathcal{D}_{{\bq}_i}) = \operatorname{softmax}(E_{{\bq}_i}\cdot E_{{\bd}_1}, \dots, E_{{\bq}_i}\cdot E_{{\bd}_{|\mathcal{D}_{{\bq}_i}|}}) \nonumber,$} \\
    \scalebox{0.875}{$P_{\text{ca}}(\cdot|{\bq}_i,\mathcal{D}_{{\bq}_i}) = \sum_{h=0}^H\sum_{t=0}^{|{\bd}_j|}\frac{\operatorname{SG}(\operatorname{CA}(0,h,t))}{H}\ |\ {\bd}_j\in\mathcal{D}_{{\bq}_i}, \nonumber$}
\end{gather}
where $\mathcal{D}_{{\bq}_i}$ is the passages returned by the dual-encoder itself and $P_{\text{ca}}$ is the target distribution gathered from the decoder's cross-attention scores. $\operatorname{SG}$ signifies stop-gradient, which prevents the decoder from being affected by the retriever loss, and $\operatorname{CA}$ denotes the cross-attention score at the last decoder layer. The term 0 refers to the first output token, $H$ is the number of cross-attention heads, and $|{\bd}_j|$ stands for the length of passage ${\bd}_j$.

The entire model is optimised to generate the target answer ${\ba}_i$ given ${\bq}_i$ and relevant passages $\mathcal{D}_{{\bq}_i}$. The final loss is: $\mathcal{L}^i_{\text{e2e}}=\mathcal{L}^i_{\text{ret}}+\mathcal{L}^i_{\text{ans}}$, where \scalebox{0.95}{$\mathcal{L}^i_{\text{ans}}=\log \prod_{t=1}^T P({\ba}^i_t|{\ba}^i_{<t},{\bq}_i, \mathcal{D}_{{\bq}_i})\nonumber$}.

%% file: sections/3_experiments.tex
\section{Experiments}

\subsection{Datasets and Metrics}
We evaluate on the \textsc{Xor-TyDi QA} dataset~\citep{asai-etal-2021-xor}, with XOR-Retrieve for cross-lingual retrieval and XOR-Full for multilingual open-retrieval QA. We conduct zero-shot evaluations on two benchmarks, MIRACL~\citep{zhang-etal-2023-miracl} for monolingual retrieval and MKQA~\citep{longpre-etal-2021-mkqa} for multilingual open-domain QA. For XOR-Retrieve, we use the February 2019 English Wikipedia dump as the retrieval corpus and the same dumps from 13 languages for XOR-Full and MKQA~\citep{asai-etal-2021-xor}. For MIRACL, we use the monolingual Wikipedia preprocessed by~\citet{zhang-etal-2023-miracl}. Following prior work, we evaluate models at Recall@5kt (top 5000 tokens) on XOR-Retrieve; F1, exact match (EM) and BLEU on XOR-Full; nDCG@10 on MIRACL; and F1 on MKQA.

\subsection{Baselines}
We compare \ours with three ranges of baselines: (i) Zero-shot baselines ("-En") fine-tuned on supervised English-only training data. We consider Natural Questions~\citep{kwiatkowski-etal-2019-natural} as the default. (ii) Supervised baselines that fine-tuned on human-annotated multilingual data (\ie \textsc{Xor-TyDi QA}). (iii) Few-shot models that improve zero-shot baselines with only a few supervised multilingual instances.

\paragraph{Retriever Baselines.}

For XOR-Retrieve, we include: (1) Zero-shot retrievers: translate-test methods: DPR+MT~\citep{asai-etal-2021-xor} and ReATT+MT~\citep{jiang-etal-2022-retrieval}; models pre-trained on multilingual Wikipedia: CLASS-En~\citep{jiang-etal-2024-pre} and LAPCA~\citep{lapca} (2) Supervised retrievers: multilingual dense retrievers: mDPR~\citep{asai-etal-2021-xor}, CORA~\citep{asai2021one}, Sentri~\citep{sorokin-etal-2022-ask}, QuiCK~\citep{ren-etal-2022-empowering}; token-level dense retrievers: DrDecr~\citep{li-etal-2022-learning-cross} pre-trains ColBERT on WikiMatrix~\citep{schwenk-etal-2021-wikimatrix}. (3) Few-shot retrievers: SWIM-X~\citep{thakur-etal-2024-leveraging} generates massive synthetic data from LLMs through a summarisation-then-ask technique. CLASS (5-shot) fine-tunes CLASS-En on our 5-shot examples. For MIRACL~\citep{zhang-etal-2023-miracl}, we include two supervised retrievers: fine-tuned mContriever~\citep{izacard2022unsupervised} and Hybrid that combines the results of BM25, mDPR, and mColbert~\citep{colbertv1}. 

\paragraph{Reader Baselines.}
(1) Zero-shot baselines: translate-test methods MT+DPR, ReAtt+MT, and GMT+GS generate answers from English retrieved passages with question and answer translations. (2) Supervised baselines: BM25 does in-language retrieval with an extractive multilingual QA model; MT+Mono first applies BM25 and then MT+DPR if no answer was generated. Fusion-in-decoder methods (\ie CORA, CLASS, Sentri, LAPCA) use retrieval-augmented generation, generating target language answers from multilingual retrieved passages. (3) Few-shot readers: Gemma (5-shot)~\citep{team2024gemma} and LLaMa3 (5-shot)~\citep{touvron2023llama} prompt LLMs with few-shot examples and retrieved passages using the template in Appendix Table~\ref{tab:llm_fs_qa_prompt}; CLASS (5-shot) fine-tunes CLASS-En on few-shot examples. We use the same 5-shot examples for all methods.

\subsection{Implementation Details}\label{sec:implementation}
With the proposed self-supervised data construction method, we generate 18,735,159 triplets for pre-training across 8 languages, with statistics in Appendix Table~\ref{tab:data_statistics}. We initialise our model from the mT5-large checkpoint~\citep{xue-etal-2021-mt5} and pre-train it using the loss function $\mathcal{L}_{\text{ssl}}$ for 100K steps with a batch size of 800 on 16 A100 GPUs for 64 hours. We set the learning rate to $5\times 10^{-5}$ with 10\% steps of warm-up, and linear decay to $0$.

With our few-shot data generation method, we obtain 1,746,156 question-answer pairs across 7 languages included in \textsc{Xor-TyDi QA} after data filtering with $\mathcal{T}_l=0.5$ and $\mathcal{T}_g=0.8$, with detailed statistics shown in Table~\ref{tab:data_statistics} in Appendix. 
For fine-tuning, we first train the pre-trained model using NQ data for 8K steps and then on \ourdata for 6K-14K steps depending upon the size of the sampled
training dataset, with the loss function $\mathcal{L}_{\text{e2e}}$. We set the batch size to 128 and the learning rate to $5\times 10^{-5}$. We apply an asynchronous passage update mechanism, where we periodically refresh the retrieved passages for each training query using the most recent checkpoint every 1K steps.

\input{tables/xor_retrieve}
\input{tables/miracl}

\subsection{Retrieval Results}
\paragraph{XOR-Retrieve.}
Table~\ref{tab:xor_retrieve_results} shows that \ours, fine-tuned on 100K synthetic data, surpasses the few-shot SWIM-X (7M) by 5.5\% at Recall@5kt, despite the latter using substantially more synthetic data generated by a significantly larger proprietary LLM (PaLM2). This indicates our method's great efficiency in training and data generation. Further scaling up the training data to full size does not improve retrieval accuracy. In addition, we find that fine-tuning CLASS, a sophisticated pre-training method, on the same set of 5-shot examples, lags \ours by 3.1 points. This shows our method of amplifying data through LLM prompting is superior to direct fine-tuning.

\paragraph{MIRACL.}
Table~\ref{tab:miracl} shows that \ours surpasses the few-shot retriever SWIM-X by 5.1\%, although SWIM-X generates synthetic data on each MIRACL language through 3-shot prompting, whereas \ours is exclusively trained on synthetic data generated from 5-shot examples of $\textsc{Xor-TyDi QA}$ and thus, evaluated on a \emph{zero-shot} manner. We further divide languages into seen and unseen groups based on \ours's training data. It outperforms SWIM-X on all seen languages and 7 out of 10 unseen languages, except on \texttt{zh}, \texttt{fr}, and \texttt{de}. We suspect SWIM-X benefits significantly from large-scale synthetic data generation on these high-resource languages.

\input{tables/xor_full}
\input{tables/mkqa}

\subsection{Multilingual QA Results}
\paragraph{XOR-Full.}
In Table~\ref{tab:xor_full_results}, we show \ours achieves the best results in few-shot settings, outperforming CLASS-En (directly fine-tuning on 5-shot examples) by 8.4\% and directly few-shot promoting LLMs for QA by 18\%. Compared to supervised readers, \ours surpasses CORA and other pipeline methods while achieving results comparable to the rest. It is also noteworthy that in two low-resource languages, \ours outperforms comparable supervised baselines in Bengali and achieves a closer match in Telugu, indicating the effectiveness of our method in handling low-resource languages.  

\paragraph{MKQA.}
In Table~\ref{tab:mkqa_qa_zs}, \ours achieves the best zero-shot results on MKQA in almost all languages, with an improvement of +2.8\% compared to supervised CORA and CLASS. This suggests that training on our synthetic data can well generalise to other new languages, indicating that generating synthetic data for each target language may not be necessary for language adaptation.

\subsection{Ablation}
We perform ablation studies to justify each of our designs, with results shown in Tables~\ref{tab:ablation_cl_queries} and~\ref{tab:ablation}.

\paragraph{Cross-lingual data improves cross-lingual ability.} 
Excluding cross-lingual synthetic training data enhances performance in answering questions that require only the retrieval of in-language passages. However, the result on questions relying on cross-lingual passage retrieval declines, reducing the overall results. This is further evidenced by retrieval results R$^\text{M}$@100, where the accuracy of finding evidence in any language (\eg English and in-language) drops, with additional support from the cross-lingual passage retrieval results.

\paragraph{Data filtering improves data quality.}
By using the raw synthetic data from LLMs without any quality control, the performance suffers in every examined language except Telugu. We suspect that the NLI model is deficient in this language.

\paragraph{Geometry sampling improves long-answer generation.}
Sampling data according to geometry distribution over answer length leads to a 0.9\% gain on average. In languages that contain a significant number of long answers (\ie \texttt{ar}, \texttt{fi}, \texttt{ja}, \texttt{ru}), geometry sampling shows gains of up to 2.7\%. Conversely, in \texttt{bn}, and \texttt{ko}, where short answers dominate, random sampling is usually better.

\paragraph{Pre-training is crucial.}
We observe extremely poor results in all languages without pre-training on our \ourptdata, primarily due to the model's low retrieval accuracy in identifying relevant passages. We believe pre-training enables the model to achieve good initial retrieval accuracy, which is essential in the subsequent fine-tuning process.
\input{tables/ablation}

\subsection{Training Data Scaling}
\begin{figure}[t]
    \centering
    \setlength{\abovecaptionskip}{-0.01cm}
    \setlength{\belowcaptionskip}{-0.2cm}
    \includegraphics[width=\linewidth]{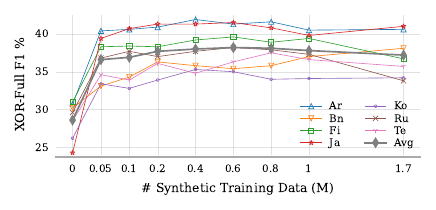}
    \caption{Performance when trained with different sizes of our synthetic data.}
    \label{fig:synthetic_data_scaling}
\end{figure}

\paragraph{Performance improves with more synthetic data.}
To investigate the effect of our data scale on models, we train \ours on subsets ranging from 0.05M to the entire 1.7M QA pairs, Results on each language and the average performance are shown in Figure~\ref{fig:synthetic_data_scaling}. As the data size increases, \ours shows enhanced average performance up to the 0.6M data scale and gradually decreases afterward. We observe that as data size increases, the proportion of examples with short answers increases (78.4\% $\rightarrow$ 95.3\%), and the result on long-answer examples drops from 18.0\% to 15.1\%, indicating overfitting to short answers.

\begin{figure}[t]
    \centering
    \setlength{\belowcaptionskip}{-0.3cm}
    \includegraphics[width=\linewidth]{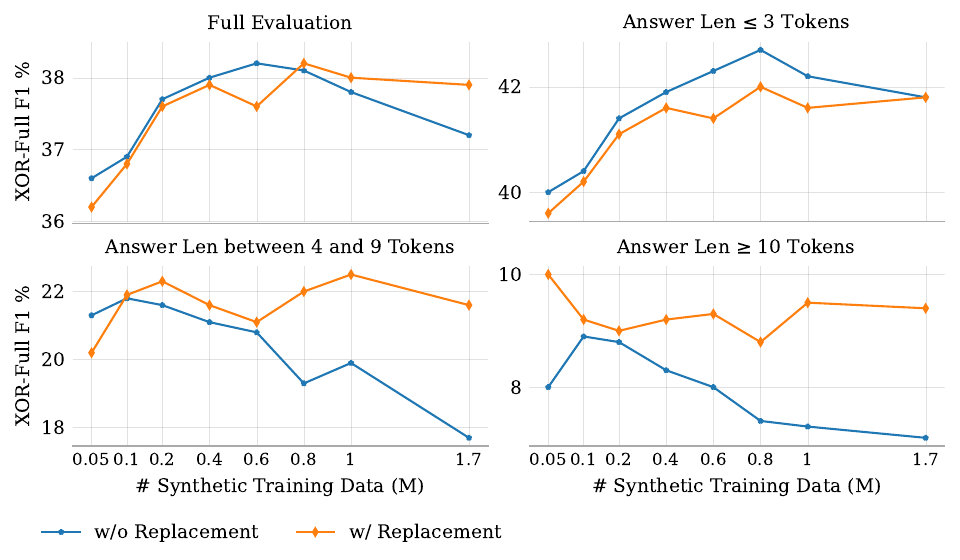}
    \caption{Performance comparison when sampling data with or without replacement by using our geometric sampling strategy.}
    \label{fig:synthetic_data_scaling_strict_geo_sampling}
\end{figure}

Our geometric sampling method (\S\ref{sec:geometric_sampling}) attempts to balance the answers by length, however its use of \emph{sampling without replacement} means the few long answer instances are quickly exhausted, such that  larger sampled datasets become skewed toward shorter answers.
To mitigate this issue, we employ \emph{sampling with replacement} variant. This method follows the precomputed geometric distribution by upsampling longer-answer examples.\footnote{We do not cap the number of repeats.} As a result, it effectively increases the number of training epochs for data points with longer answers. As shown in Figure~\ref{fig:synthetic_data_scaling_strict_geo_sampling},  \emph{sampling with replacement}  significantly improves performance on longer answers ($\ge4$ tokens) while maintaining comparable performance on shorter answers relative to the current method.



\begin{figure}[t]
    \centering
    \setlength{\abovecaptionskip}{-0.01cm}
    \setlength{\belowcaptionskip}{-0.2cm}
    \includegraphics[width=\linewidth]{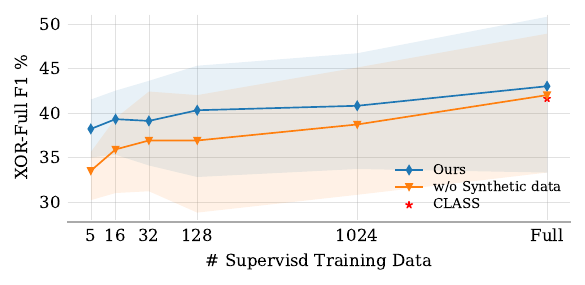}
    \caption{Results when trained with varying sizes of supervised data. The average together with the best and worst languages are reported.}
    \label{fig:supervised_data_scaling}
\end{figure}

\paragraph{Few-shot prompting is superior to direct fine-tuning and benefits from more supervised data.}
In Figure~\ref{fig:supervised_data_scaling}, we show that directly fine-tuning on the 5-shot examples is beneficial (28.6\% $\rightarrow$ 33.5\%) but remains inferior to our \emph{few-shot} method. When increasing the size of supervised data, both methods achieve consistent improvements although the performance gap narrows. With full-sized training data, \ours surpasses CLASS (43.0\% \versus 41.6\%), achieving new state-of-the-art results. See Appendix Table~\ref{tab:supervised_data_scaling_details} for results in each language.

\input{tables/prompt_source}
\subsection{Zero-shot Prompting Strategies}\label{sec:zs_cl_prompt}
We compare our \emph{few-shot} prompting strategy with two \emph{zero-shot cross-lingual} prompting methods in~\S\ref{sec:zero_shot_prompting}. In \emph{English-Prompting}, we consider NQ training data and \textsc{TyDi QA} English training data as prompting sources, respectively. In \emph{Multilingual-Prompting}, we use 5-shot examples from all languages in \textsc{Xor-TyDi QA} (\ie those used in our \emph{few-shot} setting) for prompting. When generating synthetic data for each target language, we exclude its 5-shot examples from the prompting source. We compare the success rate of generating valid examples using different prompting strategies in Appendix Table~\ref{tab:success_rate}, with \emph{few-shot} prompting achieving the highest rate and \emph{English-Prompting} with NQ yielding the lowest rate.

\paragraph{Zero-shot prompting is comparable to few-shot prompting.}
Table~\ref{tab:prompt_source_comparison_results} shows that all three zero-shot prompting variants achieve consistent improvements over \oursen with up to 8.1\% gains, highlighting the versatility of our method in zero-shot language adaptation. Prompting with English datasets created with the same guidelines achieves better results (\textsc{TyDi}-En \versus NQ-En), and using multilingual examples for prompting (\ie \textsc{Xor-TyDi}-*) is comparable to \ours. Specifically, the diversity and QA styles in prompts are more important for \texttt{fi} and \texttt{te}, while for other languages, employing in-language prompts usually leads to the best performance.

\input{tables/en_data_usage}
\paragraph{English-prompting is the best way of using English data and is complementary to existing methods.}
We compare three different ways of using \textsc{TyDi QA} English data for zero-shot learning, direct English fine-tuning, fine-tuning on machine-translated data from English, and \emph{English Prompting}. Table~\ref{tab:en_data_usage_means} shows the benefits of all three methods, with our \emph{English-Prompting} approach yielding the best results in all languages. Additionally, combining data from all three methods results in improvements over any of them when used independently, and matches the performance of our few-shot setting.

\section{Zero-shot Language Adaptation}


In~\S\ref{sec:zero_shot_prompting}, we propose a \emph{zero-shot prompting} strategy that uses few-shot examples from other languages to generate synthetic data for a distinct target language. The effectiveness of this approach is demonstrated in~\S\ref{sec:zs_cl_prompt}. In this section, we evaluate the impact of this strategy in adapting \ours to a diverse range of previously unseen languages, using only English labelled data.

\subsection{Experimental Setup}
\paragraph{Languages}
We select ten languages unseen by \ours from the MIRACL dataset for monolingual retrieval adaptation. We choose ten unseen languages from the MKQA dataset with high, medium, and low resources for multilingual open-domain QA adaptation.

\paragraph{Data Generation}
We consider the English NQ training data as the source for prompts. For each target language, we randomly sample five-shot examples from the NQ dataset to prompt the generation of Q\&A pairs from selected Wikipedia passages, following the procedure described in~\S\ref{sec:fs_data_gen}. This approach yields 128,000 training instances for each target language. Additionally, we compare this method to the translate-train baseline (MT), which uses Google Translate to translate the NQ training data into the target languages.

\paragraph{Model Training}
For both methods, we fine-tune \ours for 3K steps following the same procedure used in \ourdata (\S\ref{sec:implementation}). The final checkpoint obtained at the last training step is used for evaluation. Note that separate models are created per language in this experiment.


\input{tables/adaptation}

\subsection{Results}
\paragraph{Monolingual Retrieval Adaptation.}
As shown in the upper part of Table~\ref{tab:zs_adapt}, the zero-shot adaptation significantly improves \ours's monolingual retrieval results by an average of 5.4\% across ten unseen languages. These improvements are particularly pronounced in low-resource languages (\ie \texttt{th}, \texttt{yo}, \texttt{sw}), whereas the MT baseline results in notable declines both in these languages (\eg -36.3\% in \texttt{yo}) and overall (-3.3\%). Note that MIRACL was created by native speakers from texts in the target languages, which aligns with our data generation process. This explains the consistent gains achieved by our method and shows its superiority to translation-based approaches.

\paragraph{Multilingual Open-domain QA Adaptation.}
As shown in the bottom of Table~\ref{tab:zs_adapt}, the adaptation effectively enhances multilingual open-domain QA performance across seven languages, achieving an average improvement of 11.3\%. MT-based approaches yield results comparable to our adaptation, which is expected since MKQA was translated from NQ and the machined-translated data share the same topic distributions (\ie American-centric). In contrast, our method generates data from Wikipedia texts written in target languages to simulate how native speakers ask questions, which is more common for real-world scenarios. 

\section{Data Analysis}
\subsection{Quality Validation}
\begin{figure}[t]
    \centering
    \setlength{\abovecaptionskip}{-0.02cm}
    \setlength{\belowcaptionskip}{-0.35cm}
    \subfigure[\ourptdata (Silver-Standard)]{
        \label{fig:quality_validation_a}
        \includegraphics[width=\linewidth]{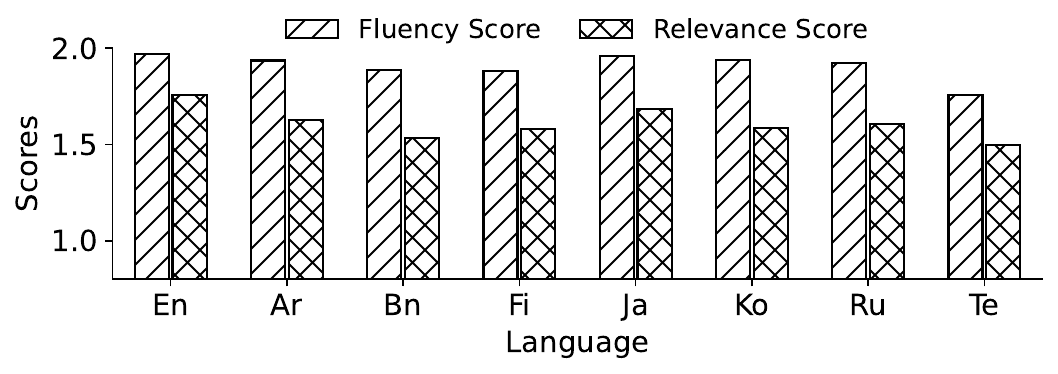}
    }
    \subfigure[\ourdata (Synthetic)]{
        \label{fig:quality_validation_b}
        \includegraphics[width=\linewidth]{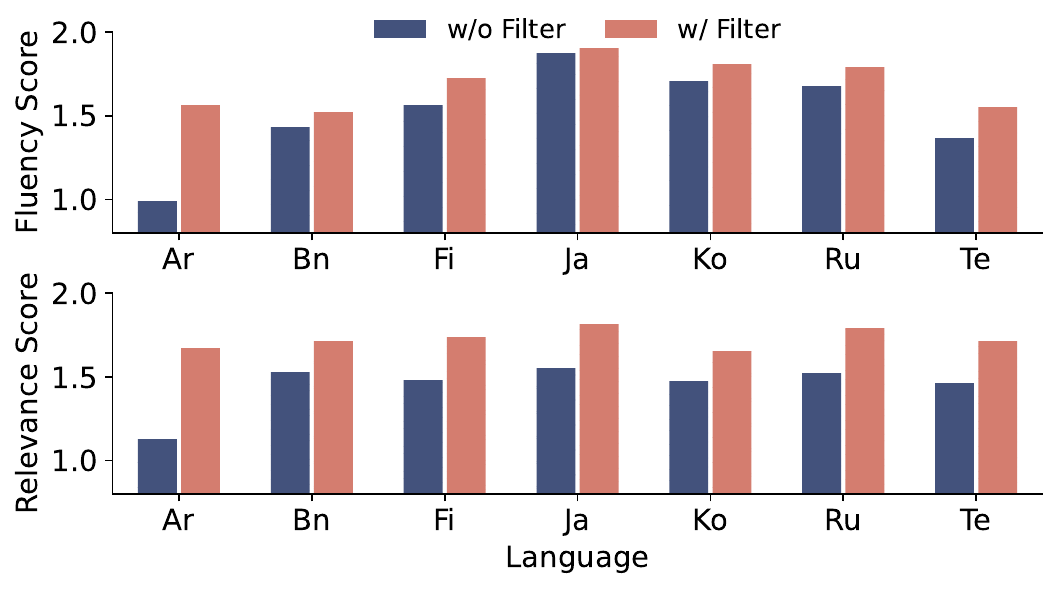}
    }
    \caption{Quality validation results on \ourptdata (top) and \ourdata (bottom). We employ \emph{Model-as-Judge} to evaluate the quality of generated data on a three-level rating scale (0-2) based on two factors: fluency and relevance.}
    \label{fig:quality_validation}
\end{figure}

\begin{figure}[t]
    \centering
    \setlength{\abovecaptionskip}{-0.01cm}
    \setlength{\belowcaptionskip}{-0.35cm}
    \includegraphics[width=\linewidth]{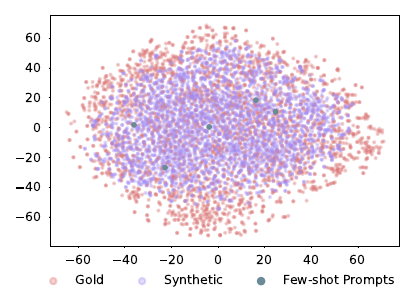}
    \caption{Distribution comparison between \ourdata and \textsc{Xor-TyDi QA} in Japanese. We show that the synthetic data is diverse and significantly overlaps with the gold standard.}
    \label{fig:tsne_gold_vs_synthetic_ja}
\end{figure}


To assess the overall quality of our synthetic data, we randomly sample 1,000 examples from the silver pre-training data (\ourptdata) and few-shot synthetic data (\ourdata). These samples are evaluated using the \texttt{GPT-4o mini} to assess quality based on: 1) Fluency (0-2): assessing whether the query is understandable, readable, and free of spelling or grammatical mistakes; 2) Relevance (0-2): evaluating the alignment between the generated query-answer pair and the passage used for data generation. The prompts employed for quality assessment are included in Appendix Table~\ref{tab:llm_assess_prompt}.

Figure~\ref{fig:quality_validation} illustrates that both types of our generated queries exhibit fluency and strong grounding in the corresponding positive passages. The silver-standard \ourptdata, derived using heuristics from WikiData  (\S\ref{sec: mlwikiqa}), consistently achieves higher scores across both metrics in all languages compared to the unfiltered synthetic \ourdata (compare Figure~\ref{fig:quality_validation}~(a) versus (b)). However, the quality of \ourdata improves significantly after applying our tailored filtering mechanism (w/ Filter columns),  almost matching the quality and fluency scores for \ourptdata. This finding underscores the critical role of the filtering procedure in producing a synthetic dataset of quality comparable to the silver-standard dataset.

\subsection{Query Distribution Comparison}

To examine the distributional differences between our synthetic \ourdata and the gold-standard data in \textsc{Xor-TyDi QA}, we randomly sample up to 20,000 examples from both datasets and visualise their distributions using t-SNE~\citep{JMLR:v9:vandermaaten08a}, which projects the queries onto a two-dimensional space. Figure~\ref{fig:tsne_gold_vs_synthetic_ja} highlights several key findings: 1) The synthetic queries exhibit sufficient diversity, as they are scattered across the plot, indicating that our approach is capable of generating queries of various types using only five labelled examples. 2) The synthetic data shows significant overlap with the gold-standard data, demonstrating that it retains the core characteristics of the gold distribution. 3) The gold-standard data exhibits greater diversity than the synthetic data, suggesting that there is still room for improvement in enhancing diversity and variation during the data generation process, which we leave for future work. Similar findings are observed in the other languages (see Appendix Figure~\ref{fig:tsne_gold_vs_synthetic}).

\subsection{Safety}
We employ \texttt{Llama-Guard-2}\footnote{\url{https://huggingface.co/meta-llama/Meta-Llama-Guard-2-8B}} as the content safety classifier to assess the presence of unsafe content within our synthetic dataset. Our analysis reveals that 98.9\% of the 1,746,156 queries in \ourdata are classified as safe.

%% file: tables/xor_retrieve.tex
\begin{table*}[t]
\setlength{\belowcaptionskip}{-0.2cm}
\setlength{\tabcolsep}{4pt}
\scriptsize
\centering
\begin{tabular}{lccccccccccc|c}
    \toprule
    \multirow{2}{*}[-1ex]{\bf Method} & \multirow{2}{*}[-1ex]{\shortstack{\bf Backbone}} & \multirow{2}{*}[-1ex]{\shortstack{\bf \# Total \\ \bf Params}} & \multirow{2}{*}[-1ex]{\shortstack{\bf Pre-training \\ \bf Data}} & \multirow{2}{*}[-1ex]{\shortstack{\bf Fine-tuning \\ \bf Data}} & \multicolumn{8}{c}{\bf R@5kt} \\
    \cmidrule(lr){6-13} 
    & & & & & Ar & Bn & Fi & Ja & Ko & Ru & Te & Avg. \\
    \midrule
    
    \multicolumn{13}{l}{\bf \emph{Zero-shot Retrievers}} \\ 
    DPR+MT$^\dagger$ & $\texttt{mBERT}$ & 220M & --- & NQ & 52.4 & 62.8 & 61.8 & 48.1 & 58.6 & 37.8 & 32.4 & 50.6 \\
    ReAtt+MT$^{\ast}$ & $\texttt{T5-L}$ & 583M & --- & NQ & 67.3 & 71.0 & 29.3 & 61.8 & 67.0 & 61.2 & 66.4 & 60.6 \\
    CLASS-En$^{\ast}$ & $\texttt{mT5-L}$ & 410M & Wikipedia & NQ & 66.7 & 78.6 & 66.6 & 60.2 & 63.2 & 58.2 & 78.2 & 67.4 \\

    \midrule
    \multicolumn{13}{l}{\bf \emph{Supervised Retrievers}} \\ 
    CORA & $\texttt{mBERT}$ & 557M & --- & NQ + XOR & 42.7 & 52.0 & 49.0 & 32.8 & 43.5 & 39.2 & 41.6 & 43.0 \\
    mDPR$^{\dagger}$ & $\texttt{mBERT}$ & 557M & --- & NQ + XOR & 48.9 & 60.2 & 59.2 & 34.9 & 49.8 & 43.0 & 55.5 & 50.2 \\
    Sentri & $\texttt{XLM-R}$ & 560M & --- & NQ + TQA + XOR & 56.8 & 62.2 & 65.5 & 53.2 & 55.5 & 52.3 & 80.3 & 60.8 \\
    QuiCK & $\texttt{mBERT}$ & 557M & --- & NQ + XOR & 63.8 & 78.0 & 65.3 & 63.5 & 69.8 & 67.1 & 74.8 & 68.9 \\
    DrDecr & $\texttt{XLM-R}$ & 278M & WikiMatrix & NQ + XOR & 70.2 & 85.9 & 69.4 & 65.1 & 68.8 & 68.8 & 83.2 & 73.1 \\
    LAPCA & $\texttt{XLM-R}$ & 560M & Wikipedia & NQ + XPAQ + XOR & 70.2 & 83.8 & 79.6 & 69.7 & 73.6 & 75.5 & 83.1 & 76.5 \\
    \class & $\texttt{mT5-L}$ & 410M & Wikipedia & NQ & 70.6 & 84.9 & 71.0 & 66.0 & 72.6 & 70.0 & 81.9 & 73.9 \\

    \midrule
    \multicolumn{13}{l}{\bf \emph{Few-shot Retrievers}} \\
    SWIM-X (7M) & $\texttt{mT5-B}$ & 580M & mC4 & SWIM-IR & 57.9 & 75.0 & 65.6 & 59.3 & 58.9 & 64.6 & 74.4 & 65.1 \\
    CLASS (5-shot) & $\texttt{mT5-L}$ & 410M & Wikipedia & NQ + XOR (5-shot) & \bf 67.0 & 78.6 & 65.6 & 59.0 & 63.6 & 59.0 & \bf 79.5 & 67.5 \\
    \rowcolor{gray!13}\bf \ours (100K) & $\texttt{mT5-L}$ & 410M & \ourptdata & NQ + \ourdata & 66.3 & 79.3 & \bf 67.8 & \bf 66.4 & 65.6 & 73.8 & 75.2 & \bf 70.6 \\
    \rowcolor{gray!23} \bf \ours (1.7M) & $\texttt{mT5-L}$ & 410M & \ourptdata & NQ + \ourdata & 63.4 & \bf 80.6 & 67.5 & 66.0 & \bf 66.7 & \bf 74.3 & 75.6 & \bf 70.6 \\

    \bottomrule
    \end{tabular}
    \caption{Results on XOR-Retrieve dev sets. Best performance is in bold. $\dagger$ and ${\ast}$ denotes results reported by~\citet{asai-etal-2021-xor} and~\citet{jiang-etal-2024-pre}, respectively. Others are copied from original papers.}
\label{tab:xor_retrieve_results}
\end{table*}

%% file: tables/miracl.tex
\begin{table*}[t]
\setlength{\belowcaptionskip}{-0.2cm}
\setlength{\tabcolsep}{3pt}
\scriptsize
\centering
\begin{tabular}{l|cccccccc|cccccccccc|c}
    \toprule
    & \multicolumn{8}{c|}{\bf Seen Languages} & \multicolumn{10}{c|}{\bf Unseen Languages} & \\
    & ar & bn & en & fi & ja & ko & ru & te & es & fa & fr & hi & id & sw & th & zh & de & yo & Avg.\\
    \midrule
    \multicolumn{20}{l}{\bf\emph{Supervised Retrievers}} \\
    Hybrid & 67.3 & 65.4 & 54.9 & 67.2 & 57.6 & 60.9 & 53.2 & 60.2 & 64.1 & 59.4  & 52.3 & 61.6 & 44.3 & 44.6 & 59.9 & 52.6 & 56.5 & 37.4 & 56.6 \\
    mContriever & 66.4 & 68.4 & 44.2 & 65.2 & 56.8 & 58.8 & 51.2 & 79.0 & 42.8 & 48.9 & 46.2 & 45.0 & 45.8 & 67.7 & 70.7 & 49.4 & 42.3 & 48.4 & 55.4 \\
    
    \midrule
    \multicolumn{20}{l}{\bf\emph{Few-shot Retrievers}} \\
    SWIM-X (180K) & 60.2 & 57.1 & 34.7 & 40.6 & 40.8 & 43.3 & 49.7 & 55.9 & 33.4 & 36.3 & \bf 64.3 & 33.0 & 39.5 & 40.0 & 56.3 & \bf 63.3 & \bf 50.2 & 36.5 & 46.4 \\
    \rowcolor{gray!23} \bf \ours (100K) & \bf 64.4 & \bf 63.6 & \bf 45.4 & \bf 64.7 & \bf 55.1 & \bf 49.6 & \bf 50.0 & \bf 76.2 & \bf 40.5 & \bf 43.7 & 36.5 & \bf 43.2 & \bf 42.6 & \bf 50.2 & \bf 60.4 & 43.2 & 36.7 & \bf 60.2 & \bf 51.5 \\
    \bottomrule
    \end{tabular}
    \caption{Monolingual retrieval results on MIRACL dev sets. Best performance is in bold. Hybrid scores are taken from~\citet{zhang-etal-2023-miracl}. mContriever and SWIM-X are copied from~\citet{thakur-etal-2024-leveraging}.}
\label{tab:miracl}
\end{table*}

%% file: tables/xor_full.tex
\begin{table*}[t]
\setlength{\belowcaptionskip}{-0.1cm}
\setlength{\tabcolsep}{2.3pt}
\scriptsize
\centering
\begin{tabular}{lccccccccccc|ccc}
    \toprule
    \multirow{2}{*}[-1ex]{\bf Method} & \multirow{2}{*}[-1ex]{\shortstack{\bf Backbone}} & \multirow{2}{*}[-1ex]{\shortstack{\bf \# Total \\ \bf Params}} & \multirow{2}{*}[-1ex]{\shortstack{\bf Pre-training \\ \bf Data}} & \multirow{2}{*}[-1ex]{\shortstack{\bf Fine-tuning \\ \bf Data}} & \multicolumn{7}{c|}{F1} & \multicolumn{3}{c}{Macro Average} \\
    \cmidrule(lr){6-12} \cmidrule(lr){13-15}
    & & & & & \bf Ar & \bf Bn & \bf Fi & \bf Ja & \bf Ko & \bf Ru & \bf Te & \bf F1 & \bf EM & \bf BLEU \\

    \midrule
    \multicolumn{15}{l}{\bf\emph{Zero-shot Readers}} \\
    MT+DPR$^\dagger$ & $\texttt{mBERT}$ & --- & --- & NQ & 7.2 & 4.3 & 17.0 & 7.9 & 7.1 & 13.6 & 0.5 & 8.2 & 3.8 & 6.8 \\
    ReAtt+MT$^\ast$ & $\texttt{T5-L}$ & 1.19B & --- & NQ & 15.0 & 10.5 & 1.8 & 13.1 & 14.9 & 15.4 & 8.2 & 11.3 & 5.5 & 9.5 \\
    GMT+GS$^\dagger$ & --- & --- & --- & NQ & 18.0 & 29.1 & 13.8 & 5.7 & 15.2 & 14.9 & 15.6 & 16.0 & 9.9 & 14.9 \\

    \midrule
    \multicolumn{15}{l}{\bf\emph{Supervised Readers}} \\
    BM25$^\dagger$ & --- & --- & --- & XOR & 31.1 & 21.9 & 21.4 & 12.4 & 12.1 & 17.7 & – & – & – & – \\
    MT+Mono$^\dagger$ & $\texttt{mBERT}$ & --- & --- & NQ + XOR & 15.8 & 9.6 & 20.5 & 12.2 & 11.4 & 16.0 & 0.5 & 17.3 & 7.5 & 10.7 \\
    CORA & $\texttt{mBERT+mT5-B}$ & 1.14B & --- & NQ + XOR & 42.9 & 26.9 & 41.4 & 36.8 & 30.4 & 33.8 & 30.9 & 34.7 & 25.8 & 23.3 \\
    \class & $\texttt{mT5-L}$ & 1.23B & Wikipedia & NQ + XOR & 49.1 &  32.0 & 46.7 & 44.1 &  38.4 & 39.9 & 41.1 & 41.6 & 32.5 & 28.2 \\
    Sentri & $\texttt{XLM-R+mT5-B}$ & 1.14B & --- & NQ + TQA + XOR & 52.5 & 31.2 & 45.5 &  44.9 & 43.1 & 41.2 & 30.7 & 41.3 & 34.9 & 30.7 \\
    LAPCA & $\texttt{XLM-R+mT5-B}$ & 1.14B & Wikipedia & NQ + XPAQ + XOR &  53.4 &  50.2 &  49.3 & 44.7 &  49.5 &  49.3 &  38.9 &  47.8 &  38.7 &  35.5 \\

    \midrule
    \multicolumn{15}{l}{\bf\emph{Few-shot Readers}} \\
    Gemma (5-shot) & $\texttt{Gemma}$ & 7B & --- & --- & 13.4 & 19.0 & 21.7 & 20.2 & 20.5 & 23.0 & 23.4 & 20.2 & 12.2 & 15.3 \\
    LLaMA3 (5-shot) & $\texttt{LLaMA3}$ & 8B & --- & --- & 22.7 & 13.2 & 22.9 & 17.8 & 19.0 & 19.2 & 28.9 & 20.5 & 12.8 & 15.6 \\
    \class (5-shot) & $\texttt{mT5-L}$ & 1.23B & Wikipedia & NQ + XOR (5-shot) & 32.3 & 28.1 & 29.9 & 25.7 & 29.5 & 27.7 & 24.7 & 29.8 & 20.5 & 21.2 \\
    \rowcolor{gray!23} \bf \ours & $\texttt{mT5-L}$ & 1.23B & \ourptdata & NQ + \ourdata & \bf 41.3 & \bf 35.4 & \bf 39.6 & \bf 41.5 & \bf 35.0 & \bf 38.2 & \bf 36.3 & \bf 38.2 & \bf 27.9 & \bf 24.4 \\
    \bottomrule

    \end{tabular}
    \caption{Multilingual QA results on the XOR-Full dev set. Best performance is in bold. $\dagger$ and $\ast$ denotes results taken from~\citet{asai2021one} and~\citet{jiang-etal-2024-pre}. Others are copied from original papers.}
\label{tab:xor_full_results}
\end{table*}

%% file: tables/mkqa.tex
\begin{table*}[t]
\setlength{\belowcaptionskip}{-0.2cm}
\setlength{\tabcolsep}{3pt}
\scriptsize
\centering
\begin{tabular}{l|ccccccccccccccccccccc}
    \toprule
    \bf Method & Da & De & Es & Fr & He & Hu & It & Km & Ms & Nl & No & Pl & Pt & Sv & Th & Tr & Vi & cn & hk & tw & Avg \\
    \midrule
    \multicolumn{22}{l}{\bf\emph{Supervised Readers}} \\
    CORA & 30.4 & 30.2 & 32.0 & 30.8 & 15.8 & 18.4 & 29.0 & 5.8 & 27.8 & 32.1 & 29.2 & 25.6 & 28.4 & 30.9 & 8.5 & 22.2 & \bf 20.9 & 5.2 & 6.7 & 5.4 & 21.8 \\
    \class & 28.3 & 32.3 & 33.3 & 31.2 & 10.3 & 23.1 & 30.6 & 7.1 & 24.7 & 30.2 & 28.4 & \bf 25.6 & 29.3 & 28.9 & 14.1 & \bf 24.8 & 19.0 & 8.0 & 7.8 & 6.7 & 22.2 \\
    
    \midrule
    \multicolumn{22}{l}{\bf\emph{Few-shot Readers}} \\
    \rowcolor{gray!23} \bf \ours & \bf 34.8 & \bf 33.3 & \bf 38.5 & \bf 34.8 & \bf 19.5 & \bf 28.4 & \bf 31.9 & \bf 7.5 & \bf 36.7 & \bf 34.1 & \bf 35.5 & 18.4 & \bf 33.4 & \bf 37.2 & \bf 15.1 & \bf 24.8 & 9.9 & \bf 9.1 & \bf 8.6 & \bf 7.9 & \bf 25.0 \\
    \bottomrule
    \end{tabular}
    \caption{Zero-shot multilingual QA results (F1) on MKQA. Best performance is in bold. "cn": "Zh-cn" (Chinese, simplified). "hk": "Zh-hk" (Chinese, Hong Kong). "tw": "Zh-tw" (Chinese, traditional). }
\label{tab:mkqa_qa_zs}
\end{table*}

%% file: tables/ablation.tex
\begin{table}[t]
\setlength{\belowcaptionskip}{-0.2cm}
\setlength{\tabcolsep}{1pt}
\scriptsize
\centering
    \begin{tabular}{l|cccc|c}
        \toprule
        & \multicolumn{4}{c|}{XOR-Full} & XOR-Retrieve \\
        \cmidrule{2-5} \cmidrule{6-6}
        & In-LG & Cross-LG & All & Retrieval & CL-Retrieval\\
        & \scriptsize Avg. F1 & \scriptsize Avg. F1 & \scriptsize Avg. F1 & \scriptsize R$^\text{M}$@100 & \scriptsize R@5kt \\
        \midrule
        \bf \ours & 46.8 & \bf 31.2 & \bf 36.9 & \bf 75.0 & \bf 70.6 \\
        \ \ - CL Queries & \bf 49.3 & 30.0 & 36.8 & 72.0 & 68.4 \\
        \bottomrule
    \end{tabular}
    \caption{The effects of generating cross-lingual queries from English passages, at 100K data scale.}
\label{tab:ablation_cl_queries}
\end{table}

\begin{table}[t]
\setlength{\belowcaptionskip}{-0.2cm}
\scriptsize
\setlength{\tabcolsep}{2.84pt}
    \begin{tabular}
    {l|cccccccc}
        \toprule
         & Ar & Bn & Fi & Ja & Ko & Ru & Te & Avg. \\
         \midrule
         \bf \ours & \bf 40.6 & 34.3 & \bf 38.4 & \bf 40.7 & 32.9 & \bf 37.7 & 33.9 & \bf 36.9 \\
         \ \ - Data Filtering & 39.0 & 31.7 & 37.4 & 39.2 & 32.3 & 35.5 & \bf 35.3 & 35.8 \\
         \ \ - Geo Sampling & 37.9 & \bf 35.9 & 36.7 & 38.5 & \bf 34.1 & 35.0 & 33.5 & 36.0 \\ 
         \ \ - \ourptdata & 11.2 & 7.2 & 10.2 & 17.5 & 7.9 & 8.5 & 4.4 & 9.6 \\
         \bottomrule
    \end{tabular}
    \caption{Ablations by removing one component of our method, at 100K data scale.}
\label{tab:ablation}
\end{table}

%% file: tables/prompt_source.tex
\begin{table}[t]
\setlength{\belowcaptionskip}{-0.2cm}
\setlength{\tabcolsep}{2.95pt}
\scriptsize
\centering
\begin{tabular}{l|cccccccc}
    \toprule
    & Ar & Bn & Fi & Ja & Ko & Ru & Te & Avg. \\
    \midrule
    \oursen & 30.7 & 30.2 & 31.0 & 24.3 & 26.2 & 29.6 & 28.5 & 28.6 \\
    \midrule
    \bf \ours & 41.7 & \bf 34.7 & 38.7 & \bf 39.4 & \bf 34.7 & \bf 35.0 & 33.5 & \bf 36.8 \\
    NQ-En & 38.8 & 33.9 & 40.1 & 33.0 & 33.0 & 34.9 & 34.3 & 35.4 \\
    \textsc{TyDi}-En & 39.1 & 34.4 & \bf 41.2 & 35.6 & 31.9 & 34.6 & \bf 36.0 & 36.1 \\
    \textsc{Xor-TyDi}-* & \bf 42.5 & 33.9 & 40.3 & 37.9 & 33.7 & 34.6 & 34.3 & 36.7 \\
    \bottomrule
    \end{tabular}
    \caption{XOR-Full performance comparison when using zero-shot prompting strategies for synthetic data generation, at 100K scale. \oursen indicates the model pre-trained on \ourptdata and fine-tuned on the English NQ dataset.}
\label{tab:prompt_source_comparison_results}
\end{table}

%% file: tables/en_data_usage.tex
\begin{table}[t]
\setlength{\belowcaptionskip}{-0.2cm}
\setlength{\tabcolsep}{2.6pt}
\scriptsize
\centering
\begin{tabular}{l|cccccccc}
    \toprule
    & Ar & Bn & Fi & Ja & Ko & Ru & Te & Avg. \\
    \midrule
    \oursen & 30.7 & 30.2 & 31.0 & 24.3 & 26.2 & 29.6 & 28.5 & 28.6 \\
    \midrule
    \ + Fine-tuning & 36.8 & 30.7 & 35.5 & 29.1 & 28.6 & 30.4 & 29.4 & 31.5 \\
    \ + Translate-train & 31.5 & 31.2 & 29.9 & 26.5 & 28.8 & 27.7 & 31.5 & 29.6 \\
    \ + English-prompt & 39.1 & 34.4 & 41.2 & 35.6 & 31.9 & 34.6 & 36.0 & 36.1 \\
    \ + All & \bf 41.9 & \bf 36.2 & \bf 43.0 & \bf 37.3 & \bf 33.7 & \bf 37.0 & \bf 37.4 & \bf 38.1 \\
    \bottomrule
    \end{tabular}
    \caption{Result comparison on XOR-Full for different means of using \textsc{TyDi}-En data.}
\label{tab:en_data_usage_means}
\end{table}

%% file: tables/adaptation.tex
\begin{table}[t]
\setlength{\belowcaptionskip}{-0.2cm}
\setlength{\tabcolsep}{1.8pt}
\scriptsize
\centering
\resizebox{\linewidth}{!}{\begin{tabular}{l|cccccccccccc}
    \toprule
   \multirow{2}{*}[-1ex]{\bf MIRACL} & \multicolumn{4}{c}{High} & \multicolumn{3}{c}{Medium} & \multicolumn{3}{c}{Low} \\
    \cmidrule(lr){2-5} \cmidrule(lr){6-8} \cmidrule(lr){9-11}
    & De & Es & Fr & Zh & Fa & Hi & Id & Sw & Th & Yo & Avg. \\
    \midrule
    \ours & 36.7 & 40.5 & 36.5 & 43.2 & 43.7 & 42.6 & 43.2 & 50.2 & 60.4 & 60.2 & 45.7 \\
    \ + MT & \bf 41.3 & \bf 41.8 & 37.1 & 41.7 & 40.7 & 42.4 & 44.1 & 50.7 & 60.5 & 23.9 & 42.4 \\
    \ + Adapt & 38.8 & 41.6 & \bf 38.6 & \bf 47.0 & \bf 47.7 & \bf 45.9 & \bf 44.2 & \bf 62.3 & \bf 66.6 & \bf 78.3 & \bf 51.1 \\
    \bottomrule
\end{tabular}}
\resizebox{\linewidth}{!}{\begin{tabular}{l|ccccccccccc}
    \toprule
   \multirow{2}{*}[-1ex]{\bf MKQA} & \multicolumn{4}{c}{High} & \multicolumn{4}{c}{Medium} & \multicolumn{2}{c}{Low} \\
    \cmidrule(lr){2-5} \cmidrule(lr){6-9} \cmidrule(lr){10-11}
    & De & Es & Fr & Zh & He & Pl & Tr & Vi & Km & Th & Avg. \\
    \midrule
    \ours & 33.3 & 38.5 & 34.8 & 8.5 & 19.5 & 18.4 & 24.9 & 9.9 & 7.5 & 15.1 & 21.0 \\
    \ + MT & \bf 42.6 & 41.6 & 41.2 & \bf 12.8 & \bf 32.1 & 29.5 & 39.9 & 39.5 & \bf 13.8 & 22.1 & 31.5 \\
    \ + Adapt & 42.1 & \bf 42.1 & \bf 43.0 & 12.0 & 27.4 & \bf 39.7 & \bf 40.4 & \bf 40.4 & 13.3 & \bf 22.7 & \bf 32.3 \\
    \bottomrule
\end{tabular}}
    \caption{Zero-shot adaptation to unseen languages in monolingual retrieval (nDCG@10) and multilingual open-domain QA (F1).}
\label{tab:zs_adapt}
\end{table}

%% file: sections/4_relatedwork.tex
\section{Related Work}
\paragraph{Pre-training for Open-domain QA.}
Open-domain QA requires retrieving relevant passages and extracting answers from them. This necessity has driven various methods that jointly train retrievers and readers. REALM~\citep{realm}, RAG~\citep{rag}, EMDR2~\citep{sachan2021endtoend}, YONO~\citep{lee-etal-2022-need}, ReAtt~\citep{jiang-etal-2022-retrieval}, and Atlas~\citep{reatt} first pre-train retrievers or initialise from pre-trained~\citep{izacard2022unsupervised} and fine-tuned retrievers. Subsequently, both components are fine-tuned jointly: the reader is trained using an answer generation loss, and the retriever is trained to promote passages that increase the likelihood of generating correct answers. Recently, this joint training mechanism has been adapted for multilingual open-domain QA~\citep{jiang-etal-2024-pre}, where retrievers are initially trained by learning from English teachers using multilingual parallel data, followed by a joint training stage with query-answer pairs generated by LLMs. Our approach follows this joint training paradigm for model pre-training but differs significantly. We use WikiData as a source to generate more informative natural questions and answers. Additionally, our pre-training method is more efficient by eliminating knowledge distillation from English models.

\paragraph{LLMs for Few-shot Data Generation.}
Prompting LLMs to generate synthetic data has been widely adopted to improve the performance of retrieval and QA tasks. UPR~\citep{sachan-etal-2022-improving} and InPars~\citep{bonifacio2022inpars} use zero-shot or few-shot prompting for passage reranking. \textsc{Promptagator}~\citep{dai2023promptagator} and SWIM-X~\citep{thakur-etal-2024-leveraging} prompt LLMs with few-shot examples to generate massive synthetic queries, either in English or in multiple languages, for retriever fine-tuning. Gecko~\citep{lee2024gecko} prompts LLMs to generate synthetic instructions and queries from Web documents and create high-quality labels for retriever fine-tuning. Beyond retrieval, LLMs are employed to generate QA data, where \textsc{QAmeleon}~\citep{qameleon} prompts a 540B LLM to generate multilingual QA pairs from only five examples. Nevertheless, these methods primarily focus on retrieval tasks and the more narrowly defined machine reading comprehension tasks. In our work, we rigorously investigate how LLMs can improve the more challenging multilingual open-domain QA tasks under few-shot settings. In addition, we explore zero-shot prompting, demonstrating that cross-lingual prompting using English data or limited multilingual data from held-out languages can yield results comparable to few-shot prompting, and we show this technique can also be leveraged for effective zero-shot language adaptation.

%% file: sections/appendix.tex


%

\begin{table*}
\propertytsize
    \centering
    \begin{tabular}{l|l|l|l}
    \toprule
    Property ID & Description & Property ID & Description \\
    \midrule
    P264 & record label & P175 & performer \\
    P176 & manufacturer & P112 & founded by \\
    P127 & owned by & P840 & narrative location \\
    P495 & country of origin & P20 & place of death \\
    P407 & language of work or name & P582 & end time \\
    P69 & educated at & P159 & headquarters location \\
    P740 & location of formation & P17 & country \\
    P136 & genre & P800 & notable work \\
    P36 & capital & P570 & date of death \\
    P190 & twinned administrative body & P4552 & mountain range \\
    P915 & filming location & P3086 & speed limit \\
    P84 & architect & P2046 & area \\
    P569 & date of birth & P86 & composer \\
    P515 & phase of matter & P2048 & height \\
    P40 & child & P580 & start time \\
    P828 & has cause & P50 & author \\
    P2067 & mass & 108 & employer \\
    P170 & creator & P2049 & width \\
    P364 & original language of film or TV show & P277 & programmed in \\
    P276 & location & P413 & position played on team / speciality \\
    P131 & located in the administrative territorial entity & P26 & spouse \\
    P106 & occupation & P607 & conflict \\
    P942 & theme music & P571 & inception \\
    P6 & head of government & P19 & place of birth \\
    P1830 & owner of & P61 & discoverer or inventor \\
    \bottomrule
    \end{tabular}
    \caption{List of English properties used for generating \ourptdata. Note that we do not generate data for a property if it does not exist in the Wikidata of target languages.}
    \label{tab:wikidata_property}
\end{table*}

\begin{table*}
\promptsize
    \centering
    \begin{tabular}{p{0.95\linewidth}}
        \toprule
        \texttt{System:} \\
        \texttt{You are a powerful question generator that generates natural and fluent questions from Wikidata triples.} \\
        \texttt{User:} \\
        \texttt{Rewrite this triplet "(\begin{CJK}{UTF8}{min}テリー・ジャックス\end{CJK}, \begin{CJK}{UTF8}{min}出生地\end{CJK}, \begin{CJK}{UTF8}{min}ウィニペグ\end{CJK})" into three questions in natural Japanese whose answer is "\begin{CJK}{UTF8}{min}ウィニペグ\end{CJK}". The questions should be diverse and the verbs used should be different. The generated questions should be separated by "\textbackslash n".} \\
        \texttt{System:} \\
        \colorbox{blue!7}{\texttt{\begin{CJK}{UTF8}{min}テリー・ジャックスの出生地はどこですか？\end{CJK}}} \\
        \colorbox{blue!7}{\texttt{\begin{CJK}{UTF8}{min}テリー・ジャックスはどこで生まれましたか?\end{CJK}}} \\
        \colorbox{blue!7}{\texttt{\begin{CJK}{UTF8}{min}テリー・ジャックスが生まれた都市は何という名前ですか？\end{CJK}}} \\
        \bottomrule
    \end{tabular}
    \begin{tabular}{p{0.95\linewidth}}
        \toprule
        \texttt{System:} \\
        \texttt{You are a powerful question generator that generates natural and fluent questions from Wikidata triples.} \\
        \texttt{User:} \\
        \texttt{Rewrite this triplet "(\begin{CJK}{UTF8}{min}レアンダー・クラブ\end{CJK}, \begin{CJK}{UTF8}{min}結成場所\end{CJK}, \begin{CJK}{UTF8}{min}ロンドン\end{CJK})" into three questions in natural Japanese whose answer is "yes". The questions should be diverse and the verbs used should be different. All elements in the triplet should be included in the question. The generated questions should be separated by "\textbackslash n".} \\
        \texttt{System:} \\
        \colorbox{blue!7}{\texttt{\begin{CJK}{UTF8}{min}レアンダー・クラブはロンドンで結成されましたか？\end{CJK}}} \\
        \colorbox{blue!7}{\texttt{\begin{CJK}{UTF8}{min}ロンドンはレアンダー・クラブの結成場所ですか？\end{CJK}}} \\
        \colorbox{blue!7}{\texttt{\begin{CJK}{UTF8}{min}レアンダー・クラブの創立がロンドンで行われたのですか？\end{CJK}}} \\
        \bottomrule
    \end{tabular}
    \caption{Examples of using ChatGPT to generate questions from triples. We use the same prompt as \texttt{Yes} questions to generate \texttt{No} ones by sampling perturbed triples. \colorbox{blue!7}{Highlighted texts} indicate system outputs.}
    \label{tab:chatgpt_prompt}
\end{table*}

\begin{table*}
\promptsize
    \centering
    \begin{tabular}{p{0.95\linewidth}}
        \toprule
        \texttt{Triple: (\begin{CJK}{UTF8}{min}伝染性単核球症\end{CJK}, \begin{CJK}{UTF8}{min}原因\end{CJK}, \begin{CJK}{UTF8}{min}エプスタイン・バール・ウイルス\end{CJK})} \\
        \texttt{Question: \begin{CJK}{UTF8}{min}どのウイルスが伝染性単核球症を引き起こすことが知られていますか？\end{CJK}} \\
        \texttt{Answer: \begin{CJK}{UTF8}{min}エプスタイン・バール・ウイルス\end{CJK}} \\
        \\ 
        \texttt{Triple: (\begin{CJK}{UTF8}{min}東日本大震災による電力危機\end{CJK}, \begin{CJK}{UTF8}{min}原因\end{CJK}, \begin{CJK}{UTF8}{min}福島第一原子力発電所事故\end{CJK})} \\
        \texttt{Question: \begin{CJK}{UTF8}{min}東日本大震災後の電力危機を引き起こした出来事は何ですか？\end{CJK}} \\
        \texttt{Answer: \begin{CJK}{UTF8}{min}福島第一原子力発電所事故\end{CJK}} \\
        \\ 
        \texttt{Triple: (\begin{CJK}{UTF8}{min}脳死\end{CJK}, \begin{CJK}{UTF8}{min}原因\end{CJK}, \begin{CJK}{UTF8}{min}脳損傷\end{CJK})} \\
        \texttt{Question: \begin{CJK}{UTF8}{min}脳死を引き起こすものとして、主に何が挙げられますか？\end{CJK}} \\
        \texttt{Answer: \begin{CJK}{UTF8}{min}脳損傷\end{CJK}} \\
        \\
        \texttt{Triple: (\begin{CJK}{UTF8}{min}壊血病\end{CJK}, \begin{CJK}{UTF8}{min}原因\end{CJK}, \begin{CJK}{UTF8}{min}ビタミンC欠乏症\end{CJK})} \\
        \texttt{Question: \colorbox{blue!7}{\begin{CJK}{UTF8}{min}何が壊血病につながるのか\end{CJK}}} \\
        \colorbox{blue!7}{\texttt{Answer: \begin{CJK}{UTF8}{min}ビタミンC欠乏症\end{CJK}}} \\
        \bottomrule
    \end{tabular}
    \caption{An example of prompting Gemma-7B to generate questions with ICL examples from ChatGPT. \colorbox{blue!7}{Highlighted texts} indicate system outputs.}
    \label{tab:icl_prompt}
\end{table*}

\begin{table*}
    \promptsize
    \centering
    \begin{tabular}{p{0.97\linewidth}}
        \toprule
        Document: \begin{CJK}{UTF8}{min}アダム・スミスは、私たちが一般的に資本主義と呼ぶものの最初の理論家と見なされています。彼の1776年の著作『国富論』は、ある安定した商業システムと評価システムの中で、個人が生産を専門化することでより多くの収益を得るというインセンティブに応えるだろうと理論化しました。これらの個人は特定の国家の介入なしに自然に「その産業を生産物が最も価値あるものとなるように指示することができる」でしょう。\end{CJK} \\
        Question: \begin{CJK}{UTF8}{min}資本主義の提唱者は誰？\end{CJK} => Answer: \begin{CJK}{UTF8}{min}アダム・スミス\end{CJK} \\
        \\
        Document: \begin{CJK}{UTF8}{min}人気のあるフロントは、1936年5月3日の総選挙で、608議席中386議席を獲得しました。初めて、社会主義者がラディカル社会主義者よりも多くの議席を獲得し、社会主義者の指導者レオン・ブルムがフランス初の社会主義首相およびその職を保持する初のユダヤ人となりました。初めての人気のあるフロント内閣は、20人の社会主義者、13人のラディカル社会主義者、2人の社会主義共和党員で構成されており（共産主義者の閣僚はいませんでした）、初めて3人の女性が含まれていました（当時、フランスでは女性が投票することができませんでした）。\end{CJK} \\
        Question: \begin{CJK}{UTF8}{min}フランスではいつから女性権利大臣が存在する？\end{CJK} => \begin{CJK}{UTF8}{min}Answer: 1936年5月3日\end{CJK} \\
        \\
        Document: \begin{CJK}{UTF8}{min}ハンガリー王国は中央ヨーロッパに存在した君主国で、中世から20世紀にかけて存在しました（1000年から1946年まで、1918年から1920年を除く）。ハンガリー公国は約1000年にエステルゴムで最初の国王スティーブン1世が戴冠し、キリスト教王国として現れました。彼の家族（アールパード王朝）は300年にわたって王国を指導しました。\end{CJK} \\
        Question: \begin{CJK}{UTF8}{min}ハンガリー王国は何年間存在した？\end{CJK} => Answer: \begin{CJK}{UTF8}{min}1000年から1946年まで、1918年から1920年を除く\end{CJK} \\
        \\
        Document: \begin{CJK}{UTF8}{min}アントマルキと英国人は別々に死体解剖報告書を書き、それぞれがナポレオンが父親を殺した病気である内出血によって死んだと結論づけました。後に、ナポレオンの髪のサンプルから高濃度のヒ素が見つかったことに基づく別の理論では、ナポレオンがヒ素中毒で死んだとされています。ただし、後の研究でもナポレオンの幼少期や息子、ジョゼフィーヌの髪のサンプルからも高濃度のヒ素が見つかりました。19世紀には医薬品やヘアクリームなどの製品でヒ素が広く使われていました。2021年に国際チームの消化器病理学者による研究では、ナポレオンは胃がんで亡くなったと結論づけられました。\end{CJK} \\
        Question: \begin{CJK}{UTF8}{min}ナポレオンが死んだのはマラリアのせい？\end{CJK} => Answer: no \\
        \\
        Document: \begin{CJK}{UTF8}{min}フィリピンの1987年憲法は次のように述べています：「国と教会の分離は不可侵であるべきです。」（第II条第6節）、および、「宗教の設立を尊重する法律は制定されず、またその自由な行使を禁止する法律も制定されません。宗教の職業と礼拝の自由な行使と享受は、差別や優遇なしに永遠に許可されます。公民権や政治的権利の行使に宗教的試験は求められません。\end{CJK} \\
        Question: \begin{CJK}{UTF8}{min}フィリピンは政教分離を原則としている？\end{CJK} => Answer: yes \\        
        \bottomrule
    \end{tabular}
    \caption{Complete prompt for few-shot question answer generation from passages in target language.}
    \label{tab:il_prompt}
\end{table*}

\begin{table*}
    \promptsize
    \centering
    \begin{tabular}{p{0.97\linewidth}}
        \toprule
        Document: Adam Smith is considered the first theorist of what we commonly refer to as capitalism. His 1776 work, An Inquiry into the Nature and Causes of the Wealth of Nations, theorized that within a given stable system of commerce and evaluation, individuals would respond to the incentive of earning more by specializing their production. These individuals would naturally, without specific state intervention, "direct ... that industry in such a manner as its produce may be of the greatest value." \\
        Question: English: Who are the proponents of capitalism? => Japanese: \begin{CJK}{UTF8}{min}資本主義の提唱者は誰\end{CJK} \\
        Answer: English: Adam Smith => Japanese: \begin{CJK}{UTF8}{min}アダム・スミス\end{CJK} \\
        \\
        Document: The Popular Front won the general election of 3 May 1936, with 386 seats out of 608. For the first time, the Socialists won more seats than the Radical-Socialists, and the Socialist leader Léon Blum became the first Socialist Prime Minister of France as well as the first Jew to hold that office. The first Popular Front cabinet consisted of 20 Socialists, 13 Radical-Socialists and two Socialist Republicans (there were no Communist Ministers) and, for the first time, included three women (women were not able to vote in France at that time). \\
        Question: English: Since when does France have a Minister of Women's Rights? => Japanese: \begin{CJK}{UTF8}{min}フランスではいつから女性権利大臣が存在する？\end{CJK} \\
        Answer: English: 3 May 1936 => Japanese: \begin{CJK}{UTF8}{min}1936 年 5 月 3 日\end{CJK} \\
        \\
        Document: The Kingdom of Hungary was a monarchy in Central Europe that existed from the Middle Ages into the twentieth century (1000–1946 with the exception of 1918–1920). The Principality of Hungary emerged as a Christian kingdom upon the coronation of the first king Stephen I at Esztergom in about the year 1000; his family (the Árpád dynasty) led the monarchy for 300 years. \\
        Question: English: How many years did the Kingdom of Hungary exist? => Japanese: \begin{CJK}{UTF8}{min}ハンガリー王国は何年間存在した\end{CJK} \\
        Answer: English: 1000 -- 1946 with the exception of 1918–1920 => Japanese: \begin{CJK}{UTF8}{min}1000年から1946年まで、1918年から1920年を除く\end{CJK} \\
        \\
        Document: Antommarchi and the British wrote separate autopsy reports, each concluding that Napoleon had died of internal bleeding caused by stomach cancer, the disease that had killed his father. A later theory, based on high concentrations of arsenic found in samples of Napoleon's hair, held that Napoleon had died of arsenic poisoning. However, subsequent studies also found high concentrations of arsenic in hair samples from Napoleon's childhood and from his son and Joséphine. Arsenic was widely used in medicines and products such as hair creams in the 19th century. A 2021 study by an international team of gastrointestinal pathologists concluded that Napoleon died of stomach cancer. \\
        Question: English: Did Napoleon die because of malaria? => Japanese: \begin{CJK}{UTF8}{min}ナポレオンが死んだのはマラリアのせい？\end{CJK}
        Answer: English: no => Japanese: no \\
        \\
        Document: The 1987 Constitution of the Philippines declares: The separation of Church and State shall be inviolable. (Article II, Section 6), and, No law shall be made respecting an establishment of religion, or prohibiting the free exercise thereof. The free exercise and enjoyment of religious profession and worship, without discrimination or preference, shall forever be allowed. No religious test shall be required for the exercise of civil or political rights. \\
        Question: English: Does the Philippines follow the principle of separation of church and state?=> Japanese: \begin{CJK}{UTF8}{min}フィリピンは政教分離を原則としている？\end{CJK} \\
        Answer: English: yes => Japanese: yes \\  
        \bottomrule
    \end{tabular}
    \caption{Complete prompt for few-shot cross-lingual question answer generation from English passages.}
    \label{tab:cl_prompt}
\end{table*}

\begin{table*}
    \promptsize
    \centering
    \begin{tabular}{p{0.97\linewidth}}
        \toprule
        English Document: Lindzen was born on February 8, 1940 in Webster, Massachusetts.[1] His father, a shoemaker, had fled Hitler's Germany with his mother. He moved to the Bronx soon after his birth and grew up in a Jewish household in a predominantly Catholic neighborhood there.[3][5] Lindzen attended the Bronx High School of Science (winning Regents' and National Merit Scholarships), Rensselaer Polytechnic Institute,and Harvard University.[6] From Harvard, he received an A.B. in physics in 1960, followed by an S.M. in applied mathematics in 1961 and a PhD in applied mathematics in 1964. His doctoral thesis, Radiative and photochemical processes in strato- and mesospheric dynamics,[7] concerned the interactions of ozone photochemistry, radiative transfer, and dynamics in the middle atmosphere. \\
        English Question: Where was Richard Siegmund Lindzen born? => English Answer: Webster, Massachusetts \\
        \\
        English Document: On 30 September 2011, Justice Johnson Lam of the Court of First Instance of the High Court (CFI) ruled in Vallejos' case that existing legislation restricting FDHs from qualifying for permanent residence contravened the Hong Kong Basic Law. Lam also found that Vallejos and Domingo, but not the three other applicants, had fulfilled the condition of taking Hong Kong as their only permanent home and being ordinarily resident in Hong Kong for seven years. The Court of Appeal of the High Court overturned the CFI's decision on Vallejos' case on 28 March 2012. Vallejos and Domingo then jointly appealed to the Court of Final Appeal (CFA), which heard their case on 26–28 February 2013; the CFA rejected their appeal on 25 March 2013. \\
        English Question: Who was the judge in the case of Vallejos and Domingo v. Commissioner of Registration? => English Answer: Justice Johnson Lam \\
        \\
        English Document: Human dissections were carried out by the Greek physicians Herophilus of Chalcedon and Erasistratus of Chios in the early part of the third century BC.[7] During this period, the first exploration into full human anatomy was performed rather than a base knowledge gained from 'problem-solution' delving.[8] While there was a deep taboo in Greek culture concerning human dissection, there was at the time a strong push by the Ptolemaic government to build Alexandria into a hub of scientific study.[8] For a time, Roman law forbade dissection and autopsy of the human body,[9] so physicians had to use other cadavers. Galen, for example, dissected the Barbary macaque and other primates, assuming their anatomy was basically the same as that of humans.[10][11][12] \\
        English Question: Who first performed human dissection? => English Answer: Herophilus of Chalcedon and Erasistratus of Chios \\
        \\
        English Document: On 16 March 1915, Watson gained his Royal Aero Club Certificate No. 1,117 (equivalent of a pilot's licence) with the London and Provincial School at the London Aerodrome, Hendon, having sought a commission with the Royal Naval Air Service with the outbreak of war in 1914.[6] Sadly, on 30 June 1915 he lost his life when the Caudron G.3 aeroplane he was flying disintegrated in flight and crashed in Dunlye field, a few miles from the Cross-in-Hand Hotel, Sussex. Watson is buried in Dundee's Western Cemetery.[2] \\
        English Question: Where is Preston Albert Watson buried? => English Answer: Dundee's Western Cemetery \\
        \\
        English Document:  A paywall is a method of restricting access to content via a paid subscription.[1][2] Beginning in the mid-2010s, newspapers started implementing paywalls on their websites as a way to increase revenue after years of decline in paid print readership and advertising revenue.[3] In academics, research papers are often subject to a paywall and are available via academic libraries that subscribe.[4][5] \\
        English Question: What is a paywall? => English Answer: a method of restricting access to content via a paid subscription \\
        \\
        Japanese Document: \begin{CJK}{UTF8}{min}モンロー郡はは1815年6月29日に、ヨーロッパ系アメリカ人によって設立された。アラバマが州に昇格する以前のこと>だった。当初の開拓者はイギリス人の子孫であり、バージニア州、ジョージア州および両カロライナ州から来ていた。アッパー・クリーク族ウィンド・クランの著名な酋長レッド・イーグル（ウィリアム・ウェザーフォードとも呼ばれた）が、クリーク戦争（1813年-1814年）の後でこの地に入り、プランテーションを造り上げた。レッド・イーグルはクリーク>族とヨーロッパ人の血を引いており、資産である奴隷を農園主や馬の飼育者に育てた。クリーク族員の大半は、1830年代にアラバマからインディアン準州（現在のオクラホマ州）に移住させられた。その後に入ってきたヨーロッパ系アメリカ人は、奴隷労働者を連れてくるか、土地を取得した後に奴隷を購入した。\end{CJK} \\
        Japanese Question: \colorbox{blue!7}{\begin{CJK}{UTF8}{min}モンロー郡はいつ設立されましたか?\end{CJK} => Japanese Answer: \begin{CJK}{UTF8}{min}1815年6月29日\end{CJK}} \\
        \bottomrule
    \end{tabular}
    \caption{An example for zero-shot English Prompting. \colorbox{blue!7}{Highlighted texts} indicate system outputs.}
    \label{tab:en_prompt}
\end{table*}

\begin{table*}
    \promptsize
    \centering
    \begin{tabular}{p{0.97\linewidth}}
        \toprule
        Finnish Document: Tel Avivissa asuu 467 875 ihmistä, jotka jakautuvat 52 000 dunamin (52 km²; 20 neliömailia) suuruiselle alueelle, mikä tuottaa väestötiheyden 7 606 ihmistä neliökilometrillä (19 699 neliömaililla). Israelin keskusviraston (CBS) mukaan vuoteen 2009 mennessä Tel Avivin väestö kasvaa vuosittain 0,5 prosentilla. Kaikkien taustojen juutalaiset muodostavat 91,8 prosenttia väestöstä, muslimit ja arabikristityt 4,2 prosenttia ja loput kuuluvat muihin ryhmiin (mukaan lukien eri kristilliset ja aasialaiset yhteisöt). Koska Tel Aviv on monikulttuurinen kaupunki, siellä puhutaan monia kieliä heprean lisäksi. Joissakin arvioissa noin 50 000 rekisteröimätöntä afrikkalaista ja aasialaista ulkomaalaistyöntekijää asuu kaupungissa. Verrattuna länsimaisiin kaupunkeihin, rikollisuus Tel Avivissa on suhteellisen vähäistä. \\
        Finnish Question: Onko Tel Aviv monikulttuurinen maa? => Finnish Answer: no \\
        \\
        Russian Document: \cyrins{Во время войны в Вооруженных Силах Соединенных Штатов служило более 16 миллионов американцев, из которых 405 399 погибли в бою, а 671 278 были ранены. Также было 130 201 американских военнопленных, из которых 116 129 вернулись домой после войны. Ключевыми гражданскими советниками президента Рузвельта были министр войны Генри Л. Стимсон, который мобилизовал промышленность и центры призыва для обеспечения армии, командованной генералом Джорджем Маршаллом, и Военно-воздушных сил под командованием генерала Хапа Арнольда. Военно-морской флот, возглавляемый министром военно-морского флота Фрэнком Ноксом и адмиралом Эрнестом Кингом, оказался более автономным. Общие приоритеты устанавливал Рузвельт.} \\
        Russian Question: \cyrins{Сколько американских солдат участвовало во Второй Мировой войне?} => Russian Answer: \cyrins{16 миллионов} \\
        \\
        Bengali Document: {\bng dikKeNr ibrued/dh sNNGgRam calaena Jara kansas-enbRas/ka Aa{I}enr pRit iberadhii icheln, es simMilt HJeichl sn 1854 et. es Aa{I}n EkiT Aa{I}n ichl Ja kansas {O} enbRas/ka Elakar pish/cmaNJ/cel mailkana kRmbr/dhn sm/bhb krt. es kLaiskYal ilbaeriljm EbNNG Ar/th{oi}nitk sNNGs/kaer smr/thn kret, ikn/tu muk/t ANJ/cel gulaimr pRshareNr ibrued/dh ichl. pair/Tr AidhkaNNGsh{I} pRathimkbhaeb dikKeN Upis/thit ichl, ikn/tu Ut/ter sphl ichl. 1858 saelr pr/Jn/t, EiT AidhkaNNGsh puur/bgamii {O} pRak/tn iphR s{I}laredr sHeJaigta inJe Ut/terr pRaJ sms/t raejY bD bD brRHt/trtW gThn kerichl. sada dikKeNr manuShra gulaimr ibpn/ntay icin/tt HJe UeThicheln. 1860 sael pRthm irpaibLkan raSh/TRpit AabRaHam ilNNGkenr inr/bacenr saeth, mHan dikKNii raSh/TRguil {I}Una{I}eTD es/TTs ethek ibic/chn/n HJe egl.} \\
        Bengali Question: {\bng mair/kn Juk/traeSh/TRr irpabilkan pair/Tr pRthm epRiseDn/T ek icheln ?} => Bengali Answer: {\bng AabRaHam ilNG/kn} \\
        \\
        Arabic Document: \aratext{الدروع، والكتيبة الأولى من الفرسان الثامنة دمرت الملاجئ العراقية والمركبات القتالية في قطاع الفرقة العراقية الخامسة والعشرين للمشاة. في 24 فبراير 1991، تقدمت الفرقة الثانية من الفرقة الأولى للمشاة عبر الثغر في الدفاع العراقي غرب وادي البطين وكذلك قامت بتطهير القطاع الشمالي الشرقي لموقع الثغر من المقاومة العدوانية. قامت مجموعة المهمة 3-37 من الدروع بكسر الدفاع العراقي وتطهير أربعة ممرات وتوسيع الفجوة تحت نيران العدو المباشرة. أيضًا، في 24 فبراير، قامت الفرقة الأولى من الفرقة الأولى للمشاة بالتعاون مع الفرقة الأولى من الفرسان بتدمير الحواجز العراقية ودورياتها التابعة للفرقة العراقية السادسة والعشرين.} \\
        Arabic Question: \aratext{ماهو اسم أول منطقة عراقية تعرضت للغزوالأمريكي؟} => Arabic Answer: \aratext{وادي الباطن} \\
        \\
        Korean Document: \begin{CJK}{UTF8}{mj}스미스-풋넘 풍력 터빈은 세계 최초의 1메가와트 규모의 풍력 터빈이었습니다. 1941년에 버몬트주 캐슬턴의 그랜드파스 노브에 연결되어 현지 전력 공급 시스템에 연결되었습니다. 이 터빈은 파머 코슬렛 풋넘이 디자인하고 S. 모건 스미스 회사에서 제조했습니다. 이 1.25메가와트 터빈은 1100시간 동안 작동한 후 전쟁 중 재료 부족으로 약점이 알려진 곳에서 날개가 파손되었습니다. 이후 1979년까지는 최대 규모의 풍력 터빈으로 남았습니다.\end{CJK} \\
        Korean Question: \begin{CJK}{UTF8}{mj}세상에서 가장 큰 풍력 에너지 발전소는 무엇인가?\end{CJK} => Korean Answer: \begin{CJK}{UTF8}{mj}스미스-퍼트남 풍력 터빈\end{CJK} \\
        \\
        Japanese Document: \begin{CJK}{UTF8}{min}細菌性髄膜炎の原因として>多い肺炎球菌と髄膜炎菌は鼻咽腔上皮細胞に付着しコロニーを形成する。そこから血管内に侵入し脳室内脈絡叢に到達する。脈絡叢上皮細胞に直接感染し脳脊髄液中に入ることができる。脳脊髄液中では免疫防御機構が機能しないため細菌は急速に増殖する。細菌性髄膜炎の発症機序において重要なのは浸潤した細菌が誘発する炎症反応である。細菌性髄膜炎の神経症状や合併症の多くは、細菌による組織の直接的な破壊よりもむしろ、浸潤した細菌に対する免疫応答によって引き起こされている。結果として、抗生物質療法により脳脊髄液が無菌化された後になっても神経の損傷は進行しうる\end{CJK} \\
        Japanese Question: \colorbox{blue!7}{\begin{CJK}{UTF8}{min}細菌性髄膜炎の最も一般的な原因は何か?\end{CJK} => Japanese Answer: \begin{CJK}{UTF8}{min}肺炎球菌と髄膜炎菌は鼻咽腔上皮細胞に付着しコロニーを\end{CJK}} \\ \colorbox{blue!7}{\begin{CJK}{UTF8}{min}形成する。\end{CJK}} \\
        \bottomrule
    \end{tabular}
    \caption{An example for zero-shot Multilingual Prompting. \colorbox{blue!7}{Highlighted texts} indicate system outputs.}
    \label{tab:multilingual_prompt}
\end{table*}

\begin{table*}
    \scriptsize
    \centering
    \begin{tabular}{p{0.97\linewidth}}
        \toprule
        \texttt{Question:} \textcolor{red}{\{Example question \#1\}} \\
        \texttt{Answer}: \textcolor{red}{\{Example answer \#1\}} \\
        $\cdots$
        \\
        \texttt{Question}: \textcolor{red}{\{Example question \#5\}} \\
        \texttt{Answer}: \textcolor{red}{\{Example answer \#5\}} \\
        \\
        \texttt{Passage \#1 Title}: \textcolor{red}{\{Passage \#1 Title\}} \\
        \texttt{Passage \#1 Text}: \textcolor{red}{\{Passage \#1 Text\}} \\
        $\cdots$
        \\
        \texttt{Passage \#N Title}: \textcolor{red}{\{Passage \#N Title\}} \\
        \texttt{Passage \#N Text}: \textcolor{red}{\{Passage \#N Text\}} \\
        \\
        \texttt{Task description: predict the} \textcolor{red}{\{Test Question Language\}} \texttt{answer to the following question. The answer should be a minimal span extracted from the document. You should only output the answer.} \\
        \\
        \texttt{Question:} \textcolor{red}{\{Test question\}} \\
        \texttt{Answer:} \\
        \bottomrule
    \end{tabular}
    \caption{Prompt template for few-shot multilingual QA with LLMs.}
    \label{tab:llm_fs_qa_prompt}
\end{table*}

\begin{table*}
    \scriptsize
    \centering
    \begin{tabular}{p{0.97\linewidth}}
        \toprule
        Relevance Assessment \\
        \midrule
        You are given a Q\&A pair and a paragraph. Your goal is to Rate the relevance of the Q\&A pair to the paragraph on a scale from 0 to 2. \\
        0: Very low relevance, the Q\&A pair and paragraph are almost unrelated. \\
        1: Moderate relevance, the Q\&A pair and paragraph share some overlap. \\
        2: High relevance, the Q\&A pair are strongly grounded by the paragraph. \\
        Output Format: \\
        Relevance (0-2) \\
        Only provide the final result in the above structured format without any additional explanations. \\
        
        Paragraph: \textcolor{red}{\{Paragraph\}} \\
        Q: \textcolor{red}{\{Synthetic Query\}} \\
        A: \textcolor{red}{\{Synthetic Answer\}} \\
        \bottomrule
        \\
        \toprule
        Fluency Assessment \\
        \midrule
        You are given a question. Your goal is to Rate the fluency of the question on a scale from 0 to 2. \\
        0: Poor fluency, the question is unclear, contains significant grammatical errors, or is incomprehensible. \\
        1: Moderate fluency, the question has minor grammatical errors or awkward phrasing but is still understandable. \\
        2: High fluency, the question is clear, well-structured, and grammatically correct. \\
        Output Format: \\
        Fluency (0-2): \\
        Only provide the final result in the above structured format without any additional explanations. \\
        
        Question: \textcolor{red}{\{Synthetic Query\}} \\
        \bottomrule
    \end{tabular}
    \caption{Prompt template for quality validation of synthetic data using \texttt{Model-as-Judge}.}
    \label{tab:llm_assess_prompt}
\end{table*}

\begin{table*}[t]
\setlength{\belowcaptionskip}{-0.3cm}
\setlength{\tabcolsep}{7pt}
\footnotesize
\centering
\begin{tabular}{l|ccc|ccc}
    \toprule
    & \multicolumn{3}{c|}{\bf\ourptdata} & \multicolumn{3}{c}{\bf\ourdata} \\
    \cmidrule(lr){2-4} \cmidrule(lr){5-7}
    & \# Q-A Paris & Question Length & Answer Length & \# Q-A Paris & Question Length & Answer Length \\
    \midrule
    \texttt{Arabic} & 1,803,765 & 7.00$_{\pm\text{2.13}}$ & 1.65$_{\pm\text{0.84}}$ & 80,575 & 8.20$_{\pm\text{2.86}}$ & 1.57$_{\pm\text{1.30}}$ \\
    \texttt{Bengali} & 407,496 & 6.13$_{\pm\text{1.80}}$ & 1.65$_{\pm\text{0.85}}$ & 127,562 & 8.97$_{\pm\text{2.99}}$ & 1.63$_{\pm\text{1.44}}$ \\
    \texttt{English} & 7,963,985 & 7.95$_{\pm\text{2.54}}$ & 1.78$_{\pm\text{1.01}}$ & --- & --- & --- \\
    \texttt{Finnish} & 2,135,790 & 6.02$_{\pm\text{1.75}}$ & 1.32$_{\pm\text{0.64}}$ & 270,627 & 5.83$_{\pm\text{2.09}}$ & 1.38$_{\pm\text{0.90}}$ \\
    \texttt{Japanese} & 2,735,635 & 14.74$_{\pm\text{3.57}}$ & 3.57$_{\pm\text{1.73}}$ & 143,265 & 10.19$_{\pm\text{2.18}}$ & 3.96$_{\pm\text{4.69}}$ \\
    \texttt{Korean} & 1,018,348 & 5.46$_{\pm\text{1.78}}$ & 1.55$_{\pm\text{0.80}}$ & 192,002 & 5.72$_{\pm\text{2.29}}$ & 1.42$_{\pm\text{0.92}}$ \\
    \texttt{Russian} & 2,561,925 & 6.94$_{\pm\text{2.17}}$ & 1.70$_{\pm\text{1.10}}$ & 792,914 & 7.34$_{\pm\text{2.64}}$ & 1.44$_{\pm\text{1.03}}$ \\
    \texttt{Telugu} & 108,215 & 5.60$_{\pm\text{1.84}}$ & 1.50$_{\pm\text{0.74}}$ & 139,211 & 6.48$_{\pm\text{2.48}}$ & 1.49$_{\pm\text{1.17}}$ \\
    \bottomrule
    \end{tabular}
    \caption{Dataset statistics of our pre-training data \ourptdata and few-shot synthetic data \ourdata in each language.}
\label{tab:data_statistics}
\end{table*}



\begin{table*}[t]
\setlength{\belowcaptionskip}{-0.3cm}
\setlength{\tabcolsep}{7pt}
\footnotesize
\centering
\begin{tabular}{l|ccccccc|c}
    \toprule
    Prompting Strategy & Ar & Bn & Fi & Ja & Ko & Ru & Te & Avg. \\
    \midrule
    \ours & 5.9\% & 13.9\% & 16.9\% & 7.7\% & 21.4\% & 15.4\% & 13.2\% & 13.4\% \\
    NQ-En & 3.4\% & 8.8\% & 8.2\% & 1.2\% & 8.3\% & 5.9\% & 4.2\% & 5.3\% \\
    \textsc{TyDi}-En & 5.0\% & 13.6\% & 12.8\% & 1.9\% & 17.0\% & 6.3\% & 5.9\% & 7.0\% \\
    \textsc{Xor-TyDi-*} & 10.2\% & 14.6\% & 14.2\% & 2.5\% & 22.7\% & 9.1\% & 10.7\% & 9.8\% \\
    \bottomrule
    \end{tabular}
    \caption{Success Rate of synthetic data generation across seven languages with different prompting strategies. Success Rate = valid examples after data filtering / total examples (\ie \# Documents)}
\label{tab:success_rate}
\end{table*}

\begin{table*}[t]
\footnotesize
\centering
\begin{tabular}{lccccccc|ccc}
    \toprule
    \multirow{2}{*}[-1ex]{\bf Method} & \multicolumn{7}{c|}{F1} & \multicolumn{3}{c}{Macro Average} \\
    \cmidrule(lr){2-8} \cmidrule(lr){9-11}
    & \bf Ar & \bf Bn & \bf Fi & \bf Ja & \bf Ko & \bf Ru & \bf Te & \bf F1 & \bf EM & \bf BLEU \\
    \midrule
    \multicolumn{11}{l}{\bf\emph{5-shot}} \\
    \oursen & 35.6 & 32.7 & 35.5 & 35.1 & 30.2 & 33.6 & 31.8 & 33.5 & 23.8 & 23.0 \\
    \ours & \bf 41.3 & \bf 35.4 & \bf 39.6 & \bf 41.5 & \bf 35.0 & \bf 38.2 & \bf 36.3 & \bf 38.2 & \bf 27.9 & \bf 24.4 \\

    \midrule
    \multicolumn{11}{l}{\bf\emph{16-shot}} \\
    \oursen & 38.3 & 31.0 & 39.4 & 38.3 & 35.2 & 34.9 & 34.6 & 35.9 & 26.1 & 24.1 \\
    \ours & \bf 42.0 & \bf 35.6 & \bf 41.4 & \bf 41.7 & \bf 35.3 & \bf 39.2 & \bf 40.0 & \bf 39.3 & \bf 29.3 & \bf 26.6 \\

    \midrule
    \multicolumn{11}{l}{\bf\emph{32-shot}} \\
    \oursen & 42.4 & 31.2 & 40.8 & 38.1 & 33.0 & 37.9 & 34.9 & 36.9 & 26.3 & 25.5 \\
    \ours & \bf 43.6 & \bf 35.6 & \bf 42.2 & \bf 42.5 & \bf 34.1 & \bf 38.6 & \bf 37.0 & \bf 39.1 & \bf 28.8 & \bf 26.6 \\

    \midrule
    \multicolumn{11}{l}{\bf\emph{128-shot}} \\
    \oursen & 42.0 & 28.8 & 41.7 & 40.3 & \bf 34.6 & 34.7 & 36.0 & 36.9 & 27.0 & 25.2 \\
    \ours & \bf 45.3 & \bf 32.8 & \bf 44.3 & \bf 43.8 & 34.0 & \bf 39.9 & \bf 42.1 & \bf 40.3 & \bf 30.5 & \bf 27.4 \\

    \midrule
    \multicolumn{11}{l}{\bf\emph{1024-shot}} \\
    \oursen & 45.0 & 30.8 & 45.1 & 39.2 & 34.1 & 39.1 & 37.5 & 38.7 & 29.3 & 26.5 \\
    \ours & \bf 47.5 & \bf 33.7 & \bf 46.7 & \bf 41.4 & \bf 35.9 & \bf 40.2 & \bf 40.1 & \bf 40.8 & \bf 31.3 & \bf 27.9 \\

    \midrule
    \multicolumn{11}{l}{\bf\emph{full}} \\
    \oursen & 48.9 & \bf 33.3 & 47.7 & 42.9 & \bf 39.6 & 40.0 & 41.7 & 42.0 & 32.7 & 28.5 \\
    \ours & \bf 50.8 & \bf 33.3 & \bf 47.8 & \bf 45.0 & 38.9 & \bf 42.0 & \bf 43.1 & \bf 43.0 & \bf 33.4 & \bf 29.6 \\

    \bottomrule
    \end{tabular}
    \caption{Detailed results in each language when trained with varying sizes of supervised data.}
\label{tab:supervised_data_scaling_details}
\end{table*}

\begin{figure*}[t]
    \centering
    \setlength{\abovecaptionskip}{-0.01cm}
    \setlength{\belowcaptionskip}{-0.35cm}
    \includegraphics[width=\linewidth]{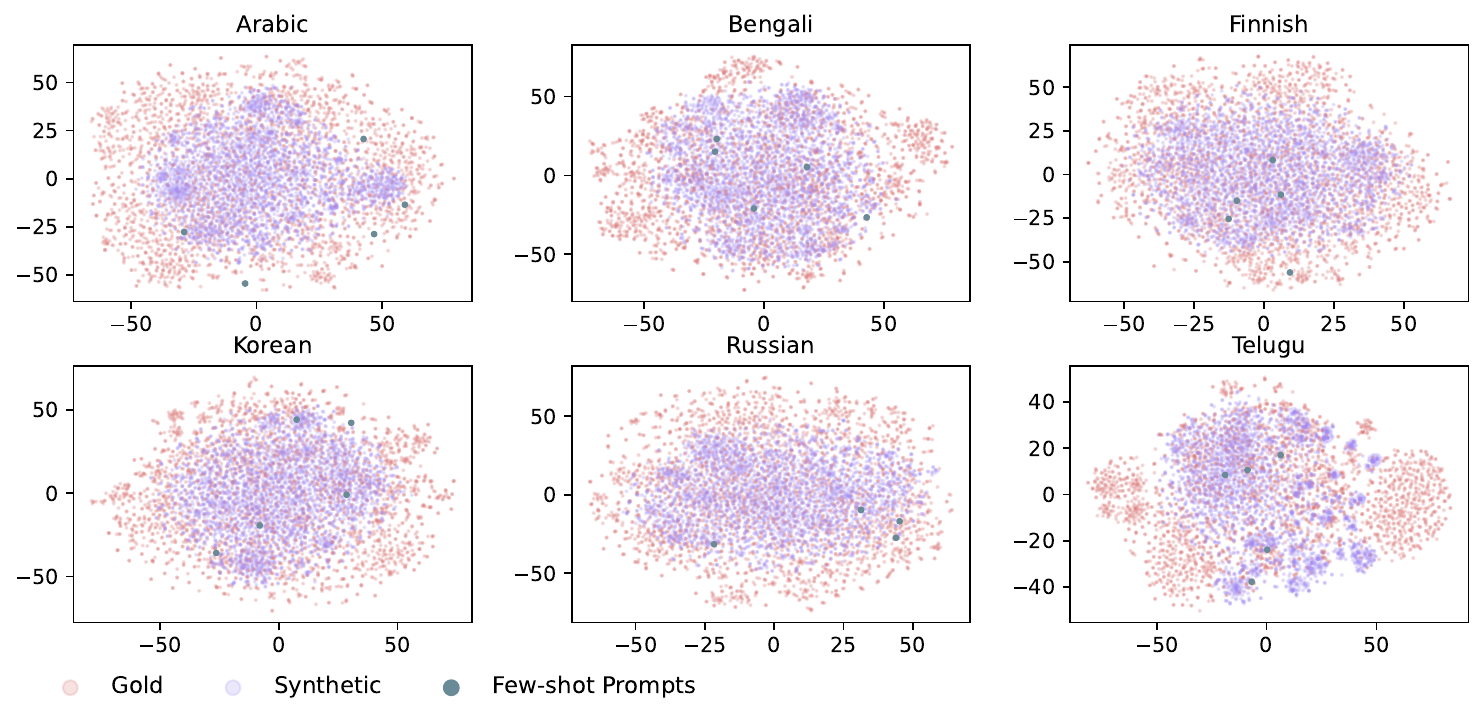}
    \caption{Distribution comparison between \ourdata and \textsc{Xor-TyDi QA} in the rest languages. We demonstrate that the diverse synthetic data can be expanded from only five-shot examples and retains the core characteristics of the gold distribution.}
    \label{fig:tsne_gold_vs_synthetic}
\end{figure*}